\documentclass[nohyperref]{article}

\usepackage{microtype}
\usepackage{booktabs} 

\usepackage[noend]{algpseudocode}
\usepackage{algorithm}
\usepackage{subcaption}
\usepackage[pdftex]{graphicx} 

\usepackage{amsmath,amsfonts,bm}

\def\eqref#1{equation~\ref{#1}}

\def\1{\bm{1}}

\DeclareMathAlphabet{\mathsfit}{\encodingdefault}{\sfdefault}{m}{sl}
\SetMathAlphabet{\mathsfit}{bold}{\encodingdefault}{\sfdefault}{bx}{n}

\newcommand{\medianS}{h}
\newcommand{\anyMedianS}{\medianS}

\usepackage{hyperref}

\usepackage[accepted]{icml2022}

\usepackage{amsmath}
\usepackage{amssymb}
\usepackage{mathtools}
\usepackage{amsthm}

\usepackage[capitalize,noabbrev]{cleveref}

\theoremstyle{plain}
\newtheorem{theorem}{Theorem}[section]

\newtheorem{corollary}[theorem]{Corollary}
\theoremstyle{definition}
\newtheorem{definition}[theorem]{Definition}

\theoremstyle{remark}

\usepackage[textsize=tiny]{todonotes}

\icmltitlerunning{Certified Neural Network Watermarks with Randomized Smoothing}

\begin{document}

\twocolumn[
\icmltitle{Certified Neural Network Watermarks with Randomized Smoothing}




\icmlsetsymbol{equal}{*}

\begin{icmlauthorlist}
\icmlauthor{Arpit Bansal}{equal,umd}
\icmlauthor{Ping-yeh Chiang}{equal,umd}
\icmlauthor{Michael Curry}{umd}
\icmlauthor{Rajiv Jain}{adobe}
\icmlauthor{Curtis Wigington}{adobe}
\icmlauthor{Varun Manjunatha}{adobe}
\icmlauthor{John P Dickerson}{umd}
\icmlauthor{Tom Goldstein}{umd}
\end{icmlauthorlist}

\icmlaffiliation{umd}{University of Maryland, College Park}
\icmlaffiliation{adobe}{Adobe Research, USA}

\icmlcorrespondingauthor{Arpit Bansal}{bansal01@umd.edu}
\icmlcorrespondingauthor{Ping-yeh Chiang}{pchiang@cs.umd.edu}

\icmlkeywords{Machine Learning, ICML}

\vskip 0.3in
]



\printAffiliationsAndNotice{\icmlEqualContribution} 

\begin{abstract}
Watermarking is a commonly used strategy to protect creators' rights to digital images, videos and audio. Recently, watermarking methods have been extended to deep learning models -- in principle, the watermark should be preserved when an adversary tries to copy the model. However, in practice, watermarks can often be removed by an intelligent adversary. Several papers have proposed watermarking methods that claim to be empirically resistant to different types of removal attacks, but these new techniques often fail in the face of new or better-tuned adversaries. In this paper, we propose a \emph{certifiable} watermarking method. Using the randomized smoothing technique proposed in Chiang et al., we show that our watermark is guaranteed to be unremovable unless the model parameters are changed by more than a certain $\ell_2$ threshold. In addition to being certifiable, our watermark is also empirically more robust compared to previous watermarking methods. Our experiments can be reproduced with code at \url{https://github.com/arpitbansal297/Certified_Watermarks}
\end{abstract}

\section{Introduction}

With the rise of deep learning, there has been an extraordinary growth in the use of neural networks in various computer vision and natural language understanding tasks. In parallel with this growth in applications, there has been exponential growth in terms of the cost required to develop and train state-of-the-art models \citep{amodei2018ai}. For example, the latest GPT-3 generative language model \citep{brown2020gpt3} is estimated to cost around $4.6$ million dollars \citep{gpt3cost} in TPU cost alone. This does not include the cost of acquiring and labeling data or paying engineers, which may be even greater. With up-front investment costs growing, if access to models is offered as a service, the incentive is strong for an adversary to try to steal the model, sidestepping the costly training process. Incentives are equally strong for companies to protect such a significant investment.

Watermarking techniques have long been used to protect the copyright of digital multimedia \citep{hartung1999wmhistory}. The copyright holder hides some imperceptible information in images, videos, or sound. When they suspect a copyright violation, the source and destination of the multimedia can be identified, enabling appropriate follow-up actions \citep{hartung1999wmhistory}. Recently, watermarking has been extended to deter the theft of machine learning models \citep{uchida2017embedding, zhang2018protecting}. The model owner either imprints a predetermined signature into the parameters of the model \citep{uchida2017embedding} or trains the model to give predetermined predictions \citep{zhang2018protecting} for a certain trigger set (e.g. images superimposed with a predetermined pattern). 

A strong watermark must also resist removal by a motivated adversary. Even though the watermarks in \citep{uchida2017embedding, zhang2018protecting, adi2018turning} initially claimed some resistance to various watermark removal attacks, it was later shown in \citep{shafieinejad2019robustness_remove, aiken2020remove} that these watermarks can in fact be removed with more sophisticated methods, using a combination of distillation, parameter regularization, and finetuning. To avoid the cat-and-mouse game of ever-stronger watermark techniques that are only later defeated by new adversaries, we propose a certifiable watermark: unless the attacker changes the model parameters by more than a certain $\ell_2$ distance, the watermark is guaranteed to remain.

To the best of our knowledge, our proposed watermarking technique is the first to provide a certificate against an $\ell_2$ adversary. We also analyzed whether $\ell_2$ adversary is a reasonable threat model as well as the magnitude of appropriate defense radius. Surprisingly, we find our certified radius to be quite substantial relative to the range of meaningful radius that one could certify. Additionally, we empirically find that our certified watermark is more resistant to previously proposed watermark removal attacks \citep{shafieinejad2019robustness_remove, aiken2020remove} compared to its counterparts -- it is thus valuable even when a certificate is not required.

\section{Related Work}

\paragraph{Watermark techniques}
\citep{uchida2017embedding} proposed the first method of watermarking neural networks: they embed the watermark into the parameters of the network during training through regularization. However, the proposed approach requires explicit inspection of the parameters for ownership verification. Later, \citep{zhang2018protecting, rouhani2018deepsigns} improved upon this approach, such that the watermark can be verified through API-only access to the model. Specifically, they embed the watermark by forcing the network to deliberately misclassify certain ``backdoor'' images. The ownership can then be verified through the adversary's API by testing its predictions on these images.

In light of later and stronger watermark removal techniques \citep{aiken2020remove, whiteboxremoval, shafieinejad2019robustness_remove}, several papers have proposed methods to improve neural network watermarking. \citep{wang2019whiteboxrobust} propose an improved white-box watermark that avoids the detection and removal techniques from \citep{whiteboxremoval}.  \citep{li2019extremevalue} propose using values outside of the range of representable images as the trigger set pattern. They show that their watermark is quite resistant to a finetuning attack. However, since their trigger set does not consist of valid images, their method does not allow for black-box ownership verification against a realistic API that only accepts actual images, while our proposed watermark is effective even in the black-box setting.

\citep{szyller2019dawn} proposed watermarking methods for models housed behind an API. Unlike our method, their method does not embed a watermark into the model weights itself, and so cannot work in scenarios where the weights of the model may be stolen directly, e.g. when the model is housed on mobile devices.

Finally, \citep{lukas2019confer} propose using a particular type of adversarial example (``conferrable'' adversarial examples) to construct the trigger set. This makes the watermark scheme resistant even to the strongest watermark removal attack: ground-up distillation which, starting from a random initialization, trains a new network to imitate the stolen model \citep{shafieinejad2019robustness_remove}. However, for their approach to be effective, they need to train a large number of models (72) on a large amount of data (e.g. requiring CINIC as opposed to CIFAR-10). While our approach does not achieve this impressive resistance to ground-up distillation, it is also much less costly.


\paragraph{Watermark removal attacks} However, one concern for all these watermark methods is that a sufficiently motivated adversary may attempt to remove the watermark. Even though \citep{zhang2018protecting, rouhani2018deepsigns, adi2018turning, uchida2017embedding} all claim that their methods are resistant to watermark removal attacks, such as finetuning, other authors  \citep{aiken2020remove, shafieinejad2019robustness_remove} later show that by adding regularization, finetuning and pruning, their watermarks can be removed without compromising the prediction accuracy of the stolen model. \cite{whiteboxremoval} shows that the watermark signals embedded by \citep{uchida2017embedding} can be easily detected and overwritten; \citep{chen2019leveragingunlabel} shows that by leveraging both labeled and unlabeled data, the watermark can be more efficiently removed without compromising the accuracy. Even if the watermark appears empirically resistant to currently known attacks, stronger attacks may eventually come along, prompting better watermark methods, and so on. 
To avoid this cycle, we propose a certifiably unremovable watermark: given that parameters are not modified more than a given threshold $\ell_2$ distance, the watermark will be preserved.

\paragraph{Certified defenses for adversarial robustness} Our work is inspired by recent work on certified adversarial robustness, \citep{cohen2019certifiedcert, chiang2019certifiedcert, wong2017provablecert, mirman2018differentiablecert, weng2018fastcert, zhang2019stablecert, eykholt2017robustcert, levine2019robustnesscert}. Certified adversarial robustness involves not only training the model to be robust to adversarial attacks under particular threat models, but also proving that no possible attacks under a particular constraint could possibly succeed. Specifically, in this paper, we used the randomized smoothing technique first developed by \citep{cohen2019certifiedcert, lecuyer2019certified} for classifiers, and later extended by \citep{chiang2020detectioncert} to deal with regression models. However, as opposed to defending against an $\ell_2$-bounded threat models in the image space, we are now defending against an $\ell_2$-bounded adversary in the parameter space.  Surprisingly, even though the certificate holds only when randomized smoothing is applied, empirically, when our watermark is evaluated in a black-box setting on the non-smoothed model, it also exhibits stronger persistence compared to previous methods.

\paragraph{Certified watermark} The only other work that we have found that proposes certified watermarks is \cite{minimal}. In \cite{minimal}, they propose a technique to find the minimal modification required to remove watermark in a neural network. Our proposal differs from theirs in two ways. First, they do not propose methods to embed a watermark that would be more resilient, rather they simply find the minimal change required to remove a watermark. On the other hand, our proposed watermark is empirically more resistant compared to previous approaches. Second, their approach is based on solving mixed integer linear programs and thus does not scale well to larger networks. For example, in their experiment, they were only able apply their technique on a network with 150 hidden neurons for MNIST \cite{minimal}. In contrast, our method can be easily applied to any modern architecture: we use ResNet-18 for all of our experiments.

\section{Methods}

Below, we introduce the formal model for neural network watermarking, and the watermark removal adversaries that we are concerned with. Then, we describe some background related to randomized smoothing, and show that by using randomized smoothing we can create a watermark that provably cannot be removed by an $\ell_2$ adversary.

\subsection{Watermarking}

\paragraph{White box vs black box}
We first introduce the distinction between black box and white box settings from the perspective of the owner of the model. In a white box setting, parameters are known. In a black box setting, the model parameters are hidden behind an API. We consider cases where the owner may have either black box or white box access to verify their watermarks.

\paragraph{Black-box watermarking}
\label{sec:backdoorsummary}
In backdoor-based watermarking, the owner employs a ``trigger set'' of specially chosen images that has disjoint distribution compared to the original dataset. If another model makes correct predictions on this trigger set, then this is evidence that the model has been stolen. A backdoor-based watermark can be verified in a black-box setting.

The trigger set may be chosen in various ways. \citep{zhang2018protecting} considered three different methods of generating the trigger set: embedded content, pre-specified noise, and abstract images. Embedded content methods embed text over existing datasets and assigns all examples with the text overlay the same fixed label. Pre-specified noise overlays Gaussian noise on top of existing dataset and again assigns the examples with the same fixed label. For abstract images, a set of images from a different domain is additionally used to train the network. For example, MNIST images could form the trigger set for a CIFAR-10 network, so if an adversary's model performs exceedingly well on MNIST images, then the adversary must have used the stolen model. Examples of trigger set images are presented can be found in Appendix - Figure \ref{fig:watermark}.

Our proposed method builds upon such backdoor-based watermarks, so our marked model can also naturally be verified in the black-box manner even though our certificate is only valid in the white-box setting described in the next section.

\paragraph{White-box watermarking}
White-box watermarks in general embed information directly into the parameters. Our proposed watermark does not directly embed information into parameters, but parameter access is required for verification, so it is still a white-box watermark. The rationale for using such a white-box watermark is detailed below. 

In the black-box setting, to verify model ownership, we generally check that the trigger set accuracy function from parameters to accuracy $f(\theta)$ is larger than a threshold \citep{shafieinejad2019robustness_remove}. The trigger set accuracy function takes in model parameter as input and outputs the accuracy on the trigger set. Since directly certifying the function is hard, we first convert the trigger set accuracy function $f(\theta)$ to its smoothed counterpart $h(\theta)$, and then check that $h(\theta)$ is greater than the threshold $t$ for ownership verification. Practically, one converts the base function to the smoothed function by injecting random noise into the parameters during multiple trigger set evaluations, and then taking the median trigger set accuracy as $\hat{h}$. Note that this verification process \emph{requires} access to parameters, so ownership verification using $\hat{h}$ is considered a \emph{white-box} watermark. 

\paragraph{Watermark Removal Threat Model}

In our experiments, we consider three different threat models to the watermark verification: 1) distillation, 2) finetuning, and 3) an $\ell_2$ adversary.

In the distillation threat model (1), we assume that the adversary initializes their model with our original model, and then trains their model with distillation using unlabeled data that comes from the same distribution. In other words, the adversary uses our original model to label the unlabeled data for finetuning. \citep{shafieinejad2019robustness_remove} propose first adding some regularization during the initial part of the attack to remove the watermark, and then later turning off the regularization to fully recover the test accuracy of the model. We experiment with this distillation attack both with and without regularization.

In the finetuning threat model (2), the adversary has its own labeled dataset from the original data-generating distribution. This adversary is strictly stronger compared to the distillation threat model. In our experiments, we make the conservative assumption that the adversary has exactly the same amount of data as the model owner.

The $\ell_2$ adversary (3) obtains the original model parameters, and then is allowed to move the parameters at most a certain $\ell_2$ distance to maximally decrease trigger set accuracy. Even though the $\ell_2$ adversary is not a completely realistic threat model, we argue similarly to the adversarial robustness literature \citep{carlini2019evaluating} that being able to defend against a small $\ell_2$ adversary is a requirement for defending against more sophisticated attacks. In our experiments, we empirically find that a large shift of parameters in $\ell_2$ distance is indicative of the strength of the adversary. For example, training the models for more time, with a larger learning rate, or using ground truth labels as opposed to distillation are all stronger attacks, and as expected, they both remove the watermark faster and move the parameters by a greater $\ell_2$ distance (Table \ref{tab:radius_l2_equiv}). Additionally, given a local Lipschitz constant of $L$ and a learning rate of $r$, the number of steps required to move outside of the $\epsilon$-$\ell_2$ ball can be upper bounded by $\epsilon/(rL)$, and we think the number of steps is a good proxy for the computational budget of the adversary.

\subsection{Watermark Certification}

For our certificates, we focus on the $\ell_2$ adversary described above: the goal of certification is to bound the worst-case decrease in trigger set accuracy, given that the model parameters do not move too far in $\ell_2$ distance. Doing this directly is in general quite difficult \citep{katz2019maraboucert}, but using techniques from \citep{chiang2020detectioncert, cohen2019certifiedcert}, we show that by adding random noise to the parameters it is possible to define a smoothed version of the model and bound the change in its trigger set accuracy.

\paragraph{Deriving the certificate}

Before we start describing the watermark certificate, we will first introduce the percentile smoothed function from \citep{chiang2020detectioncert}.

\begin{definition}\label{def:medianS}
Given $f:\mathbb{R}^d \xrightarrow{} \mathbb{R}$ and $G \sim N(0, \sigma^2 I)$, we define the {\em percentile smoothing} of $f$ as
\begin{align}
    \underline{\medianS}_{p}(x) &= \sup\{y \in \mathbb{R} \mid \mathbb{P}[f(x+G) \leq y] \leq p\} \\
    \overline{\medianS}_{p}(x) &= \inf\{y \in \mathbb{R} \mid \mathbb{P}[f(x+G) \leq y] \geq p \}
\end{align}
\end{definition}

As mentioned in \citep{chiang2020detectioncert}, the two forms $\underline{\medianS}_{p}$ and $\overline{\medianS}_{p}$ are needed to handle edge cases with discrete distributions. While $\anyMedianS_{p}$ may not admit a closed form, we can approximate it by Monte Carlo sampling~\citep{cohen2019certifiedcert}.

There are some differences from existing adversarial robustness work in how we apply these bounds. First, while the robustness literature applies the smoothing results to bound outputs of the classifier itself, we apply smoothing over the trigger set accuracy function to bound changes in trigger set accuracy. Second, we are applying smoothing over parameters as opposed to input. Our trigger set accuracy function $f(X, \theta)$ in general takes in two arguments: $X$, a set of images, and $\theta$, the model parameters. In the case of adversarial robustness, the model parameters $\theta$ are constant after training while the attacker perturbs the image $x$.  But in our case, the trigger set $X$ remains constant and the adversary can only change $\theta$. Therefore, to defend against our specific adversary, we apply smoothing over $\theta$ as opposed to $X$. Since the trigger set $X$ is constant for our case, we simply write the trigger set accuracy function as $f(\theta)$ for the remaining part of the paper. 

In our proposed watermark, we use the median smoothed version ($h_{50\%}$) of the trigger set accuracy function for ownership verification. Empirically evaluating $h_{50\%}$ essentially involves adding noise to several copies of the model parameters, calculating trigger set accuracy for all of them, and taking the median trigger set accuracy. The details of evaluating smoothed trigger set accuracy are described in Algorithm \ref{alg:pred_cert} in Appendix \ref{sec:pred_cert}.

Even though the evaluation process of $h_{50\%}$ is more involved compared to the base trigger set accuracy function, the smoothed version allows us to use Lemma 1 from \citep{chiang2020detectioncert} to bound the worst case change in the trigger set accuracy given bounded change in parameters, as shown in Corollary 1. We have delegated the proof of the corollary to Appendix \ref{sec:proof}.




\begin{corollary}
\label{eq:percentile}
Given a measureable trigger set accuracy function $f(\theta)$, the median smoothed trigger set accuracy function $\anyMedianS_{50\%}(\theta)$ can be lower bounded as follows
\begin{equation}
    \underline{\medianS}_{ \Phi \left( -\epsilon/\sigma \right)}( \theta) \ \leq \anyMedianS_{50\%}( \theta+\delta )  \quad \forall\; \|\delta\|_{2} < \epsilon,
\end{equation}
when the adversary does not modify the model parameters $\theta$ by more than $\epsilon$ in terms of $\ell_2$ norm. $\Phi$ is the standard Gaussian CDF.
\end{corollary}

Using the above corollary, we can then bound the worst case trigger set accuracy given the $\epsilon$ adversary by evaluating $\underline{\medianS}_{ \Phi \left( -\epsilon/\sigma \right)}( x) $. Even though $\underline{\medianS}_{ \Phi \left( -\epsilon/\sigma \right)}( x) $ does not have a closed form, we can calculate an empirical estimator that would lower bound it with sufficient confidence $c$. We detail steps for calculating the estimator in Algorithm \ref{alg:pred_cert} in Appendix \ref{alg:pred_cert}.

\paragraph{Trigger set accuracy and model ownership} In this paper, we assume a sufficiently high trigger set accuracy implies ownership with high probability. However, there are some scenarios where the assumption does not hold, which we will clarify below.
Whether high trigger set accuracy implies ownership depends heavily on the trigger set selected. For example, if the trigger set (X, Y) selected has labels corresponding to what most people would consider to be correct classes, then a model developed independently by someone else would likely classify such trigger sets correctly, leading to incorrect ownership assignment. However, if the trigger set consists of wrong or meaningless labels (such as dog images paired withi cat labels), then an independently developed model is very unlikely to classify such trigger sets correctly. In our paper, we assume that the trigger set examples selected have a probability less than random chance of being classified correctly by a random model, and that a sufficiently high trigger set accuracy implies ownership. Our certificate is only focused on proving the preservation of trigger set accuracy when the adversary is allowed to move parameters within a certain $\ell_2$ norm ball.

\begin{algorithm}
\begin{algorithmic}
\State \textbf{Required:} training samples $X$, trigger set samples $X_{trigger}$, learning rate $\tau$, maximum noise level $\epsilon$, replay count $k$, noise sample count $t$
\For {epoch = 1, ... , N}

\For {$B \subset X$}
\State $ g_\theta \gets E_{(x,y)\in B}[\nabla_\theta l(x, y, \theta)]$
\State $\theta \gets \theta - \tau g_\theta$ 
\EndFor
\For {$B \subset X_{trigger}$}
\State $g_\theta = 0$
\For {$i$ = 1 to $k$}
\State $\sigma \gets \frac{i}{k}\epsilon$ 
\For {$j$ = 1 to $t$}
\State $G \sim N(0, \sigma^2 I)$
\State $ g_\theta \gets g_\theta + E_{(x,y)\in B}[\nabla_\theta l(x, y, \theta+G)]$
\EndFor
\State $g_\theta \gets g_\theta/ (kt)$ 
\State $\theta \gets \theta - \tau g_\theta$ 
\EndFor

\EndFor
\EndFor

\end{algorithmic}
\caption{Embed Certifiable Watermark}
\label{alg:train}
\end{algorithm}

\paragraph{Embedding the Certifiable Watermark}

To embed the watermark during training, we add Gaussian noise and train on the trigger set images with the desired labels. For a given trigger set image, we average gradients across several (in our experiments, 100) draws of noise to better approximate the gradient of the smoothed classifier. Directly adding a large amount of noise into all parameters makes training unstable, so we incrementally increase the levels of noise within each epoch. In our experiments, we inject Gaussian noise with a range of standard deviations $\sigma$ ranging from 0 to 1. Empirically, we notice that the test accuracy drops when using this technique to embed the watermark, so to recover some of the lost test accuracy, we warm up the model with regular training and only begin embedding the watermark after the fifth epoch. We note that using warm-up epochs to recover clean accuracy is a common practice in the robustness literature \citep{balaji2019instance, gowal2018effectivenesscert}. The detailed training method is described in Algorithm \ref{alg:train}.

\section{Experiments}

In our first set of experiments, we investigate the strength of our certificate under two datasets and three watermark schemes. In our second set of experiments, we evaluate the watermark's empirical robustness to removal compared to previous methods that claimed resistance to removal attacks. The code for all these experiments is publically available \footnote{A PyTorch implementation of Certified Watermarks is available at \url{https://github.com/arpitbansal297/Certified_Watermarks}}.

\subsection{Experimental Settings}

To produce the trigger sets themselves, we consider the three schemes from \cite{zhang2018protecting}: images with embedded content (superimposed text), images with random noise, or images from an unrelated dataset (CIFAR-10 for MNIST and vice versa) (Figure \ref{fig:watermark}). While we generated certificates for all three schemes, we focus on embedded content watermark for empirical persistency evaluation.

To train the watermarked model, we used ResNet-18, SGD with learning rate of .05, momentum of .9, and weight decay of 1e-4. The model is trained for 100 epochs, and the learning rate is divided by 10 every 30 epochs. Only 50\% of the data is used for training, since we reserve the other half for the adversary. For our watermark models, we select $\sigma$ of 1, replay count of $20$, and noise sample count of $100$. Given these training parameters, embedding the watermark increase the compute time by two times compared to regular training. For certification, we use 10000 instances of Monte Carlo sampling to perform smoothing.

To attack the model, we used Adam with learning rates of .1, .001 or .0001 for 50 epochs. We test three different types of attacks: finetuning, hard-label distillation, and soft-label distillation. Soft-label distillation takes the probability distribution of the original model as labels, whereas hard-label distillation takes only the label with maximum probability. We always give the adversary the same amount of data as the owner (labeled for finetuning, unlabeled for distillation) to err on the conservative side for our evaluation.

\begin{table*}
\centering
\begin{tabular}{rccccccccc}
\toprule
Attack Radius & 0.2 &  0.4 &  0.6 &  0.8 &  1.0 &  1.2 &  1.4 &  1.6 &  1.8  \\
Worst Case Accuracy & 85.8\% &  82.5\% &  80.5\% &  76.2\% &  67.1\%  &  56.1\% & 32.0\%  &  18.4\% &  8.4\%  \\
\bottomrule
\end{tabular}
\caption{Attack Radius vs Worst Case Accuracy of the Model. It becomes meaningless to defend against a threat model with a radius larger than 1.8 because these models are indistinguishable from any randomly initialized model.}
\label{tb:appropriate_radius}
\end{table*}

\begin{figure}[htb!]
    \centering
    \includegraphics[width=0.5 \linewidth]{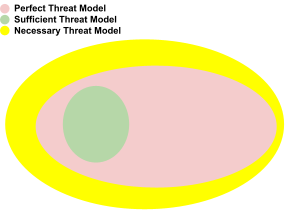}
    \caption{Graphical illustration of perfect, sufficient, and necessary threat models.}
    \label{fig:threat_model}
\end{figure}
\vspace{-2mm}

\subsection{Properties of a Good Threat Model} 

Before we present our experimental results for watermark certification, we first briefly discuss properties of a good threat model and what an appropriate radius to certify would be.

In \cite{sharif}, the author defines a \textit{perfect threat model} as being able to capture all items that are similar to the current example. In the case of adversarial attacks, we would like to capture all modified images that are similar to the original image. In the case of watermark removal, we would like to cover all attacked models that have similar test accuracy as the original model.

However, specifying a perfect threat model is often impossible for an obvious reason: we do not have an oracle measure of human perceptual similarities nor can we easily specify constraints that capture all models with similar test accuracy. As a result, researchers often have to trade off between two different imperfect threat models: a \textit{sufficient threat model} and a \textit{necessary threat model}. The sufficient threat model is a subset of the perfect threat model whereas the necessary threat model is a superset of the perfect threat model as illustrated in the Figure \ref{fig:threat_model}.

Most prior works prioritize the sufficiency criteria over the necessary criteria since being able to defend against a sufficient threat model is a requirement for being able to defend against the perfect threat model. In the watermark setting, being able to retain the trigger set accuracy (watermark) within the $\ell_2$ norm ball of the parameter space is thus a prerequisite for defending against the perfect threat model which would include models outside of the $\ell_2$ norm ball.

\paragraph{Appropriate Radius to Certify} Here, we analyze the appropriate $\ell_2$ norm constraint based on the sufficiency condition and find that a certified radius between 0 and 1.8 is appropriate and that our certified radius is indeed at a similar magnitude. In Table \ref{tb:appropriate_radius} on the right, we use a standard PGD attack in the parameter space to gauge how much the accuracy of the model could decrease within the specified radius on the CIFAR-10 dataset. We ran 40 steps of PGD with a learning rate of 0.001, and the parameter gradient is calculated over 2560 examples. If we consider all models with higher than 80\% test accuracy to be similar, then an $\ell_2$ radius of $0.6$ would satisfy the sufficiency condition. This is similar in magnitude to our certified radius presented in the next section, which is between 0.2 to 1.2. The determination of the appropriate radius is still dependent on some subjective judgement on what one would consider to be a well performing test accuracy. However, what we do know is that the appropriate radius should lie somewhere between 0 and 1.8 -- defending against a radius larger than 1.8 is meaningless, as a model with only 8\% accuracy on CIFAR-10 is indistinguishable from any randomly initialized model. 

\subsection{Watermark Certificate Evaluation}

In this section, we investigate the certified trigger set accuracy that our watermarking is able to guarantee against $\ell_2$ adversaries of various strengths. To further contextualize the meaning of a certified $\ell_2$ radius, we consider the size of the empirical changes in parameters observed after performing various watermark removal attacks. Finally, we also study how one can increase the certificate by modifying the noise level.
\vspace{-0.5mm}

As shown in Table \ref{tab:cert}, we are able to certify trigger set accuracy for radii up to 0.4 for all datasets and watermark schemes considered. This is quite a substantial radius when considering the sufficiency condition, which suggests a meaningful certificate does not exceed a radius of 1.8. Our certificate seems to be similarly effective across all trigger set types. In the best scenario for CIFAR-10, we can certify that the trigger set accuracy does not drop below 51\% as long as parameters do not move more than an $\ell_2$ distance of 1.
\vspace{-0.5mm}

\begin{table*}[htb]
\centering
\begin{tabular}{llrrrrrr}
\toprule
         &              & \multicolumn{6}{l}{$\ell_2$ Radius ($\epsilon$)} \\
Dataset  & Watermark    & 0.2    & 0.4    & 0.6   & 0.8  & 1  & 1.2   \\
\midrule 
MNIST    & Embedded content & 100\%  & 95\%   & 47\%  & 3\%  & 0\% & 0\% \\
MNIST    & Noise & 100\%  & 91\%   & 7\%  & 0\%  & 0\%& 0\%  \\
MNIST    & Unrelated & 100\%  & 94\%   & 45\%  & 4\%  & 0\% & 0\% \\
CIFAR-10 & Embedded content & 100\%  & 100\%  & 100\%  & 93\% & 51\% & 5\% \\
CIFAR-10 & Noise & 100\%  & 100\%   & 100\%  & 100\%  & 47\% & 0\% \\
CIFAR-10 & Unrelated & 100\%  & 100\%   & 100\%  & 97\%  & 35\% & 0\% \\
\bottomrule

\end{tabular}
\caption{Certified trigger set accuracy at different radius}
\label{tab:cert}
\vspace{-2mm}
\end{table*}

\begin{table*}[htb]
\centering
\begin{tabular}{llrrrrrrr}
\toprule 
&                             & \multicolumn{5}{l}{Certified Watermark Accuracy} & \multicolumn{1}{l}{} & \multicolumn{1}{l}{} \\ &
   &
  \multicolumn{1}{l}{$\ell_2$ radius ($\epsilon$)} &
  \multicolumn{1}{l}{} &
  \multicolumn{1}{l}{} &
  \multicolumn{1}{l}{} &
  \multicolumn{1}{l}{} &
  \multicolumn{1}{l}{} &
  \multicolumn{1}{l}{} \\
Noise Level ($\sigma$)          & Test Accuracy      & 0.2        & 0.4       & 0.6       & 0.8       & 1        & 1.2                  & 1.4                  \\
\midrule \\
\multicolumn{1}{r}{1}    & \multicolumn{1}{r}{86.00\%} & 100.00\%   & 100.00\%  & 100.00\%  & 93.00\%   & 51.00\%  & 5.00\%               & 0.00\%               \\
\multicolumn{1}{r}{1.1}  & \multicolumn{1}{r}{84.56\%} & 100.00\%   & 100.00\%  & 100.00\%  & 97.00\%   & 63.00\%  & 13.00\%              & 0.00\%               \\
\multicolumn{1}{r}{1.2}  & \multicolumn{1}{r}{84.18\%} & 100.00\%   & 100.00\%  & 100.00\%  & 100.00\%  & 98.00\%  & 74.00\%              & 24.00\%    \\
\bottomrule
\end{tabular}
\caption{Trade-off between certified trigger set accuracy and noise level ($\sigma$) for CIFAR-10}
\label{tab:tradeoff}
\vspace{-3mm}
\end{table*}

To see how long our certificates can persist in the face of attack, we measure the approximate amount of $\ell_2$ parameter change in the first epoch under different attack settings. In Table \ref{tab:radius_l2_equiv}, with learning rate 0.0001, parameters change by $\ell_2$ distance of approximately 2-3. In other words, it would require approximately 1/3 to 1/2 of an epoch to move outside of a certified radius of 1. (We focus here on the first epoch because changes are relatively small in succeeding epochs; see Appendix.)

Interestingly, attacks considered to be stronger correspond to changes of a greater distance. This relationship helps support the use of $\ell_2$ radius as a proxy for the strength of the adversary. For example, fine-tuning has been found to be a stronger attack compared to hard label distillation, and correspondingly \citep{shafieinejad2019robustness_remove}, fine-tuning moves the network by a larger distance in the first epoch compared to hard label distillation. Similarly, an attack that is stronger due to a higher learning rate moves the parameters much faster compared to an attack with a lower learning rate.

In Table \ref{tab:tradeoff}, we show that one can obtain larger certificates by increasing the noise level. However, as one makes the model more robust against watermark removal, the model’s test accuracy also decreases. This trade-off is similar to the trade-off observed in the adversarial robustness literature \cite{madry2017towards}. As the level of noise increases, training also becomes more unstable. For example, using the same hyperparameters as our other experiments, we were unable to train models with $\sigma = 1.5$. However, this is not to say that it is impossible to train a model with $\sigma = 1.5$. We did find an alternative setting where $\sigma = 1.5$ is trainable and offers higher robustness compared to $\sigma = 1.0-1.2$. However, since the hyperparameters are not the same, we do not list the results here as we don't think they are directly comparable.
\vspace{-1mm}

\begin{table*}[htb]
\centering
\begin{tabular}{lrrrrrr}
\toprule
Attack Type &
  Finetuning &
  \begin{tabular}[c]{@{}l@{}}Distillation \\ Hard Label\end{tabular} &
  \begin{tabular}[c]{@{}l@{}}Distillation \\ Soft Label\end{tabular} &
  Finetuning &
  \begin{tabular}[c]{@{}l@{}}Distillation \\ Hard Label\end{tabular} &
  \begin{tabular}[c]{@{}l@{}}Distillation \\ Soft Label\end{tabular} \\
Learning Rate &
  \multicolumn{1}{r}{0.0001} &
  \multicolumn{1}{r}{0.0001} &
  \multicolumn{1}{r}{0.0001} &
  \multicolumn{1}{r}{0.001} &
  \multicolumn{1}{r}{0.001} &
  \multicolumn{1}{r}{0.001} \\
  \midrule
MNIST &
  2.67 & 
  2.39 &
  1.56 &		
  19.39 &
  17.58 &
  20.35 \\	
CIFAR-10 &
  2.85 & 
  2.41	&
  2.06 &
  19.93 &
  19.40 &
  19.29  \\
  \bottomrule
\end{tabular}
\caption{$\ell_2$ distance change in the first epoch}
\label{tab:radius_l2_equiv}
\end{table*}

\begin{table*}[tb]
\centering
\begin{tabular}{llrrrr}
\toprule
Dataset &
  Attack &
  \multicolumn{1}{l}{lr} &
  \multicolumn{1}{l}{\begin{tabular}[c]{@{}l@{}}Baseline \\ Watermark\end{tabular}} &
  \multicolumn{1}{l}{\begin{tabular}[c]{@{}l@{}}Black-box \\ Watermark\end{tabular}} &
  \multicolumn{1}{l}{\begin{tabular}[c]{@{}l@{}}White-box \\ Watermark\end{tabular}} \\
  \midrule
MNIST    & Finetuning                    & 0.0001 & 45.31\% & 59.38\% & 100.00\% \\
MNIST    & Finetuning                    & 0.001  & 50.00\% & 54.70\% & 100.00\% \\
MNIST    & Hard-Label Distillation       & 0.001  & 42.19\%              & 50.00\%              & 100.00\%             \\
MNIST    & Soft-Label Distillation       & 0.001  & 96.88\%              & 100.00\%             & 100.00\%             \\
CIFAR-10 & Finetuning                    & 0.0001 & 17.20\%              & 9.40\%               & 100.00\%             \\
CIFAR-10 & Finetuning                    & 0.001  & 14.06\%              & 10.94\%              & 100.00\%             \\
CIFAR-10 & Hard-Label Distillation       & 0.001  & 29.69\%              & 81.25\%              & 100.00\%             \\
CIFAR-10 & Soft-Label Distillation       & 0.001  & 81.25\%              & 100.00\%             & 100.00\%            \\
CIFAR-100 & Finetuning                    & 0.0001  & 18.75\%              & 23.44\%              & 100.00\%             \\
CIFAR-100 & Finetuning                    & 0.001  & 0.00\%              & 0.00\%              & 0.00\%             \\
CIFAR-100 & Hard-Label Distillation       & 0.001  & 7.81\%              & 12.5\%              & 5.00\%             \\
CIFAR-100 & Soft-Label Distillation       & 0.001  & 96.88\%              & 96.88\%             & 98.44\%            \\
\midrule
MNIST    & Hard-Label Distillation + Reg & 0.1    & 40.63\%              & 32.81\%              & 0.00\%               \\
CIFAR-10 & Hard-Label Distillation + Reg & 0.1    & 8.00\%               & 27.00\%              & 0.00\%               \\
CIFAR-100 & Hard-Label Distillation + Reg & 0.1    & 0.00\%               & 0.00\%              & 0.00\%               \\
\bottomrule
\end{tabular}
\caption{Trigger set accuracy after 50 epochs of removal attacks. We note that this is only a snapshot of the trigger set accuracy. During training, trigger set accuracies could sometimes fluctuate significantly (see figures in Appendix). We use watermarks from \cite{zhang2018protecting} as the baseline watermark.}
\label{tab:attack}
\vspace{-3mm}
\end{table*}

Overall, it would take approximately 0.03 to 0.3 epochs for the attacker to escape the certified radius, depending on the type of attack, watermark schemes, and dataset. Our certified bounds are substantial when considering the sufficiency criteria, but they are still quite small when compared to a non-$\ell_2$ bounded attack. In the next section, we show that even though our certificates are not large when considering the optimization trajectory, the watermarks are empirically stronger than the certificate is able to guarantee: in most cases, our watermarks are more resistant to removal attacks compared to previous methods in both the white-box and black-box settings.

\subsection{Empirical Watermark Persistence Evaluation}

In this section, we evaluate the persistence of our proposed watermarking methods and the model's performance on the original dataset. For all experiments in this section, we use the embedded content method to produce the trigger set. We compare our watermark method with the baseline method from \cite{zhang2018protecting}, which is the same as our watermark method but without noise injection during training. We further conduct additional attack evaluations in Appendix \ref{ap:persistence}.





For persistence evaluation, we focus on two main attacks: the distillation attack and the finetuning attack, as both of these have been shown to be very effective in \citep{shafieinejad2019robustness_remove, aiken2020remove}. In addition, we tested the effect of different learning rates and label smoothing levels, which have also been shown to influence the effectiveness of watermark removal techniques \cite{shafieinejad2019robustness_remove}. To make our attacks more similar to \cite{shafieinejad2019robustness_remove}, we also experimented with adding parameter regularization during attack.

We first evaluate our proposed watermark against finetuning attacks. In Table \ref{tab:attack}, we see that our proposed watermark is much more robust with respect to finetuning attacks than the baseline method on CIFAR-10, and is comparably resistant on MNIST. In the case of CIFAR-10, the baseline watermark is completely removed within less than 10 epochs (See Figure 1 in Appendix), but our white-box watermark is still visible after finetuning for up to 50 epochs. In the case of MNIST, both the proposed method and the baseline are quite resistant. However, our proposed method achieves slightly higher trigger set accuracy for both white-box watermarks and black-box watermarks throughout the 50 epochs of the finetuning attack. In the case of CIFAR-100, neither watermark is very resistant to removal. However, our blackbox watermark slightly outperforms the baseline method.

In the face of the distillation attack, we find our white-box watermark to be extremely resistant. The trigger set accuracy remains 100\% even after 50 epochs of attack. However, our black-box watermark works more effectively on CIFAR-10 than MNIST. In the case of CIFAR-10, the black-box watermark remains at 81.25\% after 50 epochs of distillation attack, whereas only 50.00\% of trigger set accuracy remains for MNIST. In the case of CIFAR-100, our proposed watermark slightly outperforms the baseline method. However, they are both quite susceptible to removal attack.

When regularization is added in addition to distillation, we find that our white-box watermark is completely removed. This could be due to regularization moving the parameters further in terms of $\ell_2$ norm. However, we note that our black-box watermark still persists similarly to the baseline.

In some cases, the baseline watermark persists quite strongly. For example, in the case of soft-label distillation, the baseline watermark still achieves higher than 75\% accuracy after attack. We tried a variety of settings, but we had difficulty completely removing the watermark as described in \citep{shafieinejad2019robustness_remove}. Differences in performance could be due to architecture, regularization, or other factors -- experimental code was not released by \citep{shafieinejad2019robustness_remove}, so it is hard to know exactly what might be the cause. However, we note that our main goal is to show that our proposed watermark is more resistant to removal, and our trigger set accuracy is consistently higher compared to the baseline throughout the attack.

Even though our watermark is generally more resistant in both the white-box and black-box settings, our proposed method does slightly decrease the accuracy of the model on the original dataset. Test accuracies are decreased by 0.1\% (from 99.5\% to 99.4\%), 3.3\% (from 89.3\% to 86.0\%), 1.1\% (from 68.28\% to 67.23\%) for MNIST, CIFAR-10, and CIFAR-100 respectively. The decrease in clean accuracy has been historically observed for other forms of robust training \citep{madry2017towards}, and recovery of the test accuracy in robust training is still an active area of research \citep{balaji2019instance}. However, it is worth noting that the decrease in accuracy does not scale with the difficulty of the dataset. For example, even though CIFAR-100 is a much more challenging dataset compared to CIFAR-10, 
we actually observe smaller accuracy decrease for CIFAR-100.

\section{Conclusion}

We present a certifiable neural network watermark -- trigger set accuracy is provably maintained unless the network parameters are moved by more than a given $\ell_2$ distance. We see this as the first step towards guaranteed persistence of watermarks in the face of adversaries -- a valuable property in real-world applications. We also analyzed the size of our certificate with respect to the sufficiency criteria, and found that our certificates are indeed quite meaningful.

At the same time, we find that our certifiable watermarks are empirically far more resistant to removal than the certified bounds can guarantee. Indeed in the face of the removal attacks from the literature, our watermarks are more persistent than previous methods. Our randomized-smoothing-based training scheme is therefore a watermarking technique of interest even where a certificate is not needed. We are hopeful that our technique represents a contribution to both the theory and practice of neural network watermarking, and that this approach can lead to watermarks that are both empirically useful while coming with provable guarantees.

\section{Acknowledgements}
This work was supported by DARPA GARD, Adobe Research, and the Office of Naval research.

\bibliography{example_paper}

\begin{thebibliography}{34}
\providecommand{\natexlab}[1]{#1}
\providecommand{\url}[1]{\texttt{#1}}
\expandafter\ifx\csname urlstyle\endcsname\relax
  \providecommand{\doi}[1]{doi: #1}\else
  \providecommand{\doi}{doi: \begingroup \urlstyle{rm}\Url}\fi

\bibitem[Adi et~al.(2018)Adi, Baum, Cisse, Pinkas, and Keshet]{adi2018turning}
Adi, Y., Baum, C., Cisse, M., Pinkas, B., and Keshet, J.
\newblock Turning your weakness into a strength: Watermarking deep neural
  networks by backdooring.
\newblock In \emph{27th $\{$USENIX$\}$ Security Symposium ($\{$USENIX$\}$
  Security 18)}, pp.\  1615--1631, 2018.

\bibitem[Aiken et~al.(2020)Aiken, Kim, and Woo]{aiken2020remove}
Aiken, W., Kim, H., and Woo, S.
\newblock Neural network laundering: Removing black-box backdoor watermarks
  from deep neural networks.
\newblock \emph{arXiv preprint arXiv:2004.11368}, 2020.

\bibitem[Amodei \& Hernandez(2018)Amodei and Hernandez]{amodei2018ai}
Amodei, D. and Hernandez, D.
\newblock Ai and compute.
\newblock \emph{Heruntergeladen von https://blog. openai. com/aiand-compute},
  2018.

\bibitem[Balaji et~al.(2019)Balaji, Goldstein, and Hoffman]{balaji2019instance}
Balaji, Y., Goldstein, T., and Hoffman, J.
\newblock Instance adaptive adversarial training: Improved accuracy tradeoffs
  in neural nets.
\newblock \emph{arXiv preprint arXiv:1910.08051}, 2019.

\bibitem[Brown et~al.(2020)Brown, Mann, Ryder, Subbiah, Kaplan, Dhariwal,
  Neelakantan, Shyam, Sastry, Askell, et~al.]{brown2020gpt3}
Brown, T.~B., Mann, B., Ryder, N., Subbiah, M., Kaplan, J., Dhariwal, P.,
  Neelakantan, A., Shyam, P., Sastry, G., Askell, A., et~al.
\newblock Language models are few-shot learners.
\newblock \emph{arXiv preprint arXiv:2005.14165}, 2020.

\bibitem[Carlini et~al.(2019)Carlini, Athalye, Papernot, Brendel, Rauber,
  Tsipras, Goodfellow, Madry, and Kurakin]{carlini2019evaluating}
Carlini, N., Athalye, A., Papernot, N., Brendel, W., Rauber, J., Tsipras, D.,
  Goodfellow, I., Madry, A., and Kurakin, A.
\newblock On evaluating adversarial robustness, 2019.

\bibitem[Chen et~al.(2019)Chen, Wang, Ding, Bender, Jia, Li, and
  Song]{chen2019leveragingunlabel}
Chen, X., Wang, W., Ding, Y., Bender, C., Jia, R., Li, B., and Song, D.
\newblock Leveraging unlabeled data for watermark removal of deep neural
  networks.
\newblock In \emph{ICML workshop on Security and Privacy of Machine Learning},
  2019.

\bibitem[Chiang et~al.(2019)Chiang, Ni, Abdelkader, Zhu, Studor, and
  Goldstein]{chiang2019certifiedcert}
Chiang, P.-y., Ni, R., Abdelkader, A., Zhu, C., Studor, C., and Goldstein, T.
\newblock Certified defenses for adversarial patches.
\newblock In \emph{International Conference on Learning Representations}, 2019.

\bibitem[Chiang et~al.(2020)Chiang, Curry, Abdelkader, Kumar, Dickerson, and
  Goldstein]{chiang2020detectioncert}
Chiang, P.-y., Curry, M.~J., Abdelkader, A., Kumar, A., Dickerson, J., and
  Goldstein, T.
\newblock Detection as regression: Certified object detection by median
  smoothing.
\newblock \emph{arXiv preprint arXiv:2007.03730}, 2020.

\bibitem[Cohen et~al.(2019)Cohen, Rosenfeld, and
  Kolter]{cohen2019certifiedcert}
Cohen, J.~M., Rosenfeld, E., and Kolter, J.~Z.
\newblock Certified adversarial robustness via randomized smoothing.
\newblock \emph{arXiv preprint arXiv:1902.02918}, 2019.

\bibitem[Eykholt et~al.(2017)Eykholt, Evtimov, Fernandes, Li, Rahmati, Xiao,
  Prakash, Kohno, and Song]{eykholt2017robustcert}
Eykholt, K., Evtimov, I., Fernandes, E., Li, B., Rahmati, A., Xiao, C.,
  Prakash, A., Kohno, T., and Song, D.
\newblock Robust physical-world attacks on deep learning models.
\newblock \emph{arXiv preprint arXiv:1707.08945}, 2017.

\bibitem[Goldberger et~al.(2020)Goldberger, Katz, Adi, and Keshet]{minimal}
Goldberger, B., Katz, G., Adi, Y., and Keshet, J.
\newblock Minimal modifications of deep neural networks using verification.
\newblock In Albert, E. and Kovacs, L. (eds.), \emph{LPAR23. LPAR-23: 23rd
  International Conference on Logic for Programming, Artificial Intelligence
  and Reasoning}, volume~73 of \emph{EPiC Series in Computing}, pp.\  260--278.
  EasyChair, 2020.
\newblock \doi{10.29007/699q}.
\newblock URL \url{https://easychair.org/publications/paper/CWhF}.

\bibitem[Gowal et~al.(2018)Gowal, Dvijotham, Stanforth, Bunel, Qin, Uesato,
  Mann, and Kohli]{gowal2018effectivenesscert}
Gowal, S., Dvijotham, K., Stanforth, R., Bunel, R., Qin, C., Uesato, J., Mann,
  T., and Kohli, P.
\newblock On the effectiveness of interval bound propagation for training
  verifiably robust models.
\newblock \emph{arXiv preprint arXiv:1810.12715}, 2018.

\bibitem[Hartung \& Kutter(1999)Hartung and Kutter]{hartung1999wmhistory}
Hartung, F. and Kutter, M.
\newblock Multimedia watermarking techniques.
\newblock \emph{Proceedings of the IEEE}, 87\penalty0 (7):\penalty0 1079--1107,
  1999.

\bibitem[Katz et~al.(2019)Katz, Huang, Ibeling, Julian, Lazarus, Lim, Shah,
  Thakoor, Wu, Zelji{\'c}, et~al.]{katz2019maraboucert}
Katz, G., Huang, D.~A., Ibeling, D., Julian, K., Lazarus, C., Lim, R., Shah,
  P., Thakoor, S., Wu, H., Zelji{\'c}, A., et~al.
\newblock The marabou framework for verification and analysis of deep neural
  networks.
\newblock In \emph{International Conference on Computer Aided Verification},
  pp.\  443--452. Springer, 2019.

\bibitem[Lecuyer et~al.(2019)Lecuyer, Atlidakis, Geambasu, Hsu, and
  Jana]{lecuyer2019certified}
Lecuyer, M., Atlidakis, V., Geambasu, R., Hsu, D., and Jana, S.
\newblock Certified robustness to adversarial examples with differential
  privacy.
\newblock In \emph{2019 IEEE Symposium on Security and Privacy (SP)}, pp.\
  656--672. IEEE, 2019.

\bibitem[Levine \& Feizi(2019)Levine and Feizi]{levine2019robustnesscert}
Levine, A. and Feizi, S.
\newblock Robustness certificates for sparse adversarial attacks by randomized
  ablation.
\newblock \emph{arXiv preprint arXiv:1911.09272}, 2019.

\bibitem[Li(2020)]{gpt3cost}
Li, C.
\newblock \emph{OpenAI's GPT-3 Language Model: A Technical Overview}, 2020.
\newblock URL \url{https://lambdalabs.com/blog/demystifying-gpt-3/#1}.

\bibitem[Li et~al.(2019)Li, Willson, Zheng, and Zhao]{li2019extremevalue}
Li, H., Willson, E., Zheng, H., and Zhao, B.~Y.
\newblock Persistent and unforgeable watermarks for deep neural networks.
\newblock \emph{arXiv preprint arXiv:1910.01226}, 2019.

\bibitem[Lukas et~al.(2019)Lukas, Zhang, and Kerschbaum]{lukas2019confer}
Lukas, N., Zhang, Y., and Kerschbaum, F.
\newblock Deep neural network fingerprinting by conferrable adversarial
  examples.
\newblock \emph{arXiv preprint arXiv:1912.00888}, 2019.

\bibitem[Lukas et~al.(2021)Lukas, Jiang, Li, and Kerschbaum]{lukas2021sok}
Lukas, N., Jiang, E., Li, X., and Kerschbaum, F.
\newblock Sok: How robust is image classification deep neural network
  watermarking?(extended version).
\newblock \emph{arXiv preprint arXiv:2108.04974}, 2021.

\bibitem[Madry et~al.(2017)Madry, Makelov, Schmidt, Tsipras, and
  Vladu]{madry2017towards}
Madry, A., Makelov, A., Schmidt, L., Tsipras, D., and Vladu, A.
\newblock Towards deep learning models resistant to adversarial attacks.
\newblock \emph{arXiv preprint arXiv:1706.06083}, 2017.

\bibitem[Mirman et~al.(2018)Mirman, Gehr, and
  Vechev]{mirman2018differentiablecert}
Mirman, M., Gehr, T., and Vechev, M.
\newblock Differentiable abstract interpretation for provably robust neural
  networks.
\newblock In \emph{International Conference on Machine Learning}, pp.\
  3575--3583, 2018.

\bibitem[Rouhani et~al.(2018)Rouhani, Chen, and
  Koushanfar]{rouhani2018deepsigns}
Rouhani, B.~D., Chen, H., and Koushanfar, F.
\newblock Deepsigns: A generic watermarking framework for ip protection of deep
  learning models.
\newblock \emph{arXiv preprint arXiv:1804.00750}, 2018.

\bibitem[Shafieinejad et~al.(2019)Shafieinejad, Wang, Lukas, Li, and
  Kerschbaum]{shafieinejad2019robustness_remove}
Shafieinejad, M., Wang, J., Lukas, N., Li, X., and Kerschbaum, F.
\newblock On the robustness of the backdoor-based watermarking in deep neural
  networks.
\newblock \emph{arXiv preprint arXiv:1906.07745}, 2019.

\bibitem[Sharif et~al.(2018)Sharif, Bhagavatula, Bauer, and Reiter]{sharif}
Sharif, M., Bhagavatula, S., Bauer, L., and Reiter, M.~K.
\newblock Adversarial generative nets: Neural network attacks on
  state-of-the-art face recognition.
\newblock \emph{CoRR}, abs/1801.00349, 2018.
\newblock URL \url{http://arxiv.org/abs/1801.00349}.

\bibitem[Szyller et~al.(2019)Szyller, Atli, Marchal, and
  Asokan]{szyller2019dawn}
Szyller, S., Atli, B.~G., Marchal, S., and Asokan, N.
\newblock Dawn: Dynamic adversarial watermarking of neural networks.
\newblock \emph{arXiv preprint arXiv:1906.00830}, 2019.

\bibitem[Uchida et~al.(2017)Uchida, Nagai, Sakazawa, and
  Satoh]{uchida2017embedding}
Uchida, Y., Nagai, Y., Sakazawa, S., and Satoh, S.
\newblock Embedding watermarks into deep neural networks.
\newblock In \emph{Proceedings of the 2017 ACM on International Conference on
  Multimedia Retrieval}, pp.\  269--277, 2017.

\bibitem[Wang \& Kerschbaum(2019)Wang and Kerschbaum]{wang2019whiteboxrobust}
Wang, T. and Kerschbaum, F.
\newblock Robust and undetectable white-box watermarks for deep neural
  networks.
\newblock \emph{arXiv preprint arXiv:1910.14268}, 2019.

\bibitem[{Wang} \& {Kerschbaum}(2019){Wang} and {Kerschbaum}]{whiteboxremoval}
{Wang}, T. and {Kerschbaum}, F.
\newblock Attacks on digital watermarks for deep neural networks.
\newblock In \emph{ICASSP 2019 - 2019 IEEE International Conference on
  Acoustics, Speech and Signal Processing (ICASSP)}, pp.\  2622--2626, 2019.

\bibitem[Weng et~al.(2018)Weng, Zhang, Chen, Song, Hsieh, Boning, Dhillon, and
  Daniel]{weng2018fastcert}
Weng, T.-W., Zhang, H., Chen, H., Song, Z., Hsieh, C.-J., Boning, D., Dhillon,
  I.~S., and Daniel, L.
\newblock Towards fast computation of certified robustness for relu networks,
  2018.

\bibitem[Wong \& Kolter(2017)Wong and Kolter]{wong2017provablecert}
Wong, E. and Kolter, J.~Z.
\newblock Provable defenses against adversarial examples via the convex outer
  adversarial polytope.
\newblock \emph{arXiv preprint arXiv:1711.00851}, 2017.

\bibitem[Zhang et~al.(2019)Zhang, Chen, Xiao, Li, Boning, and
  Hsieh]{zhang2019stablecert}
Zhang, H., Chen, H., Xiao, C., Li, B., Boning, D., and Hsieh, C.-J.
\newblock Towards stable and efficient training of verifiably robust neural
  networks, 2019.

\bibitem[Zhang et~al.(2018)Zhang, Gu, Jang, Wu, Stoecklin, Huang, and
  Molloy]{zhang2018protecting}
Zhang, J., Gu, Z., Jang, J., Wu, H., Stoecklin, M.~P., Huang, H., and Molloy,
  I.
\newblock Protecting intellectual property of deep neural networks with
  watermarking.
\newblock In \emph{Proceedings of the 2018 on Asia Conference on Computer and
  Communications Security}, pp.\  159--172, 2018.

\end{thebibliography}
\bibliographystyle{icml2022}

\newpage
\appendix
\onecolumn
\section{Appendix}
\subsection{Algorithm for evaluating the smoothed model}
\label{sec:pred_cert}
\begin{algorithm*}
\begin{algorithmic}
\Function{TriggerSetAccuracy}{$f$, $\theta$, $\sigma$, $n$}
\State $\hat{w} \gets $ AddGaussianNoise($\theta$, $\sigma$, $n$) \Comment{$n$ simulations of noised parameter $w$}
\State $\hat{a} \gets $ $f(\hat{\theta})$ 
\Comment{evaluate trigger accuracy for each simulation of $w$}
\State $\hat{a} \gets $ Sort($\hat{a}$)
\Comment{Sort simulated accuracies}
\State $a_{median} \gets \hat{a}_{\lfloor 0.5n \rfloor}$ \Comment{Take the median}
\State \textbf{return} $a_{median}$
\EndFunction
\Function{TriggerSetAccuracyLowerBound}{$f$, $\theta$, $\sigma$, $\epsilon$, $n$, $c$}
\State $\hat{\theta} \gets $ AddGaussianNoise($\theta$, $\sigma$, $n$) \Comment{$n$ simulations of noised parameter $w$}
\State $\hat{a} \gets $ $f(\hat{\theta})$ 
\Comment{evaluate trigger accuracy for each simulation of $\theta$}
\State $\hat{a} \gets $ Sort($\hat{a}$)
\Comment{Sort simulated accuracies}
\State $k \gets$ EmpiricalPercentile($n$, $c$, $\sigma$, $\epsilon$)
\Comment{Algorithm 1 in Appendix}
\State $\underline{a} \gets \hat{a}_{k}$ 
\Comment{$\hat{a}_{k}$ Lower bound $\underline{\medianS}_{ \Phi \left( -\epsilon/\sigma \right)}( \theta)$ with confidence $c$}
\State \textbf{return} $\underline{a}$
\EndFunction

\end{algorithmic}
\caption{Evaluate and Certify the Median Smoothed Model}
\label{alg:pred_cert}
\end{algorithm*}
\subsection{Samples of watermark images}
\begin{figure}[htb]
    \centering
    
    \begin{subfigure}[t]{0.23\textwidth}
         \centering
         \includegraphics[width=1\linewidth]{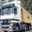}
         \caption{Original}
     \end{subfigure}
     \hfill
    \begin{subfigure}[t]{0.23\textwidth}
         \centering
         \includegraphics[width=1\linewidth]{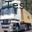}
         \caption{Embedded Content}
     \end{subfigure}
     \hfill
    \begin{subfigure}[t]{0.23\textwidth}
         \centering
         \includegraphics[width=1\linewidth]{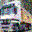}
         \caption{Gaussian Noise}
     \end{subfigure}
     \hfill
    \begin{subfigure}[t]{0.23\textwidth}
         \centering
         \includegraphics[width=1\linewidth]{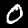}
         \caption{Unrelated}
     \end{subfigure}
     \caption{Samples of the backdoor images used for watermarking.}
     \label{fig:watermark}
\end{figure}


\clearpage
\subsection{Proof of Corollary 1}
\label{sec:proof}
\begin{align*}
    & \underline{\medianS}_{\underline{p}}( x) \ \leq \anyMedianS_{p}( x+\delta ) \ \leq \overline{\medianS}_{\overline{p}}( x) \quad && \forall\; \|\delta\|_{2} < \epsilon \;\;\; \text{Lemma 1 from \cite{chiang2020detectioncert}} \\
    => & \underline{\medianS}_{\underline{p}}( x) \ \leq \anyMedianS_{p}( x+\delta ) \quad && \forall\; \|\delta\|_{2} < \epsilon \\
    => & \underline{\medianS}_{\Phi \left( \Phi ^{-1}( p) -\frac{\epsilon}{\sigma} \right)}( x) \ \leq \anyMedianS_{p}( x+\delta ) \quad && \forall\; \|\delta\|_{2} < \epsilon  \;\;\; \text{Definition of }\underline{p} \\
    => & \underline{\medianS}_{\Phi \left( \Phi ^{-1}(50\%) -\frac{\epsilon}{\sigma} \right)}( x) \ \leq \anyMedianS_{50\%}( x+\delta ) \quad && \forall\; \|\delta\|_{2} < \epsilon  \;\;\; \text{Plug in 50\% for  }p \\
    => & \underline{\medianS}_{\Phi \left( -\frac{\epsilon}{\sigma} \right)}( x) \ \leq \anyMedianS_{50\%}( x+\delta ) \quad && \forall\; \|\delta\|_{2} < \epsilon  \;\;\; 
\end{align*}
\subsection{Trigger set trajectories during attack}
\begin{figure*}[htb]
    \centering
    \begin{subfigure}[t]{0.45\textwidth}
         \centering
         \includegraphics[width=1\linewidth]{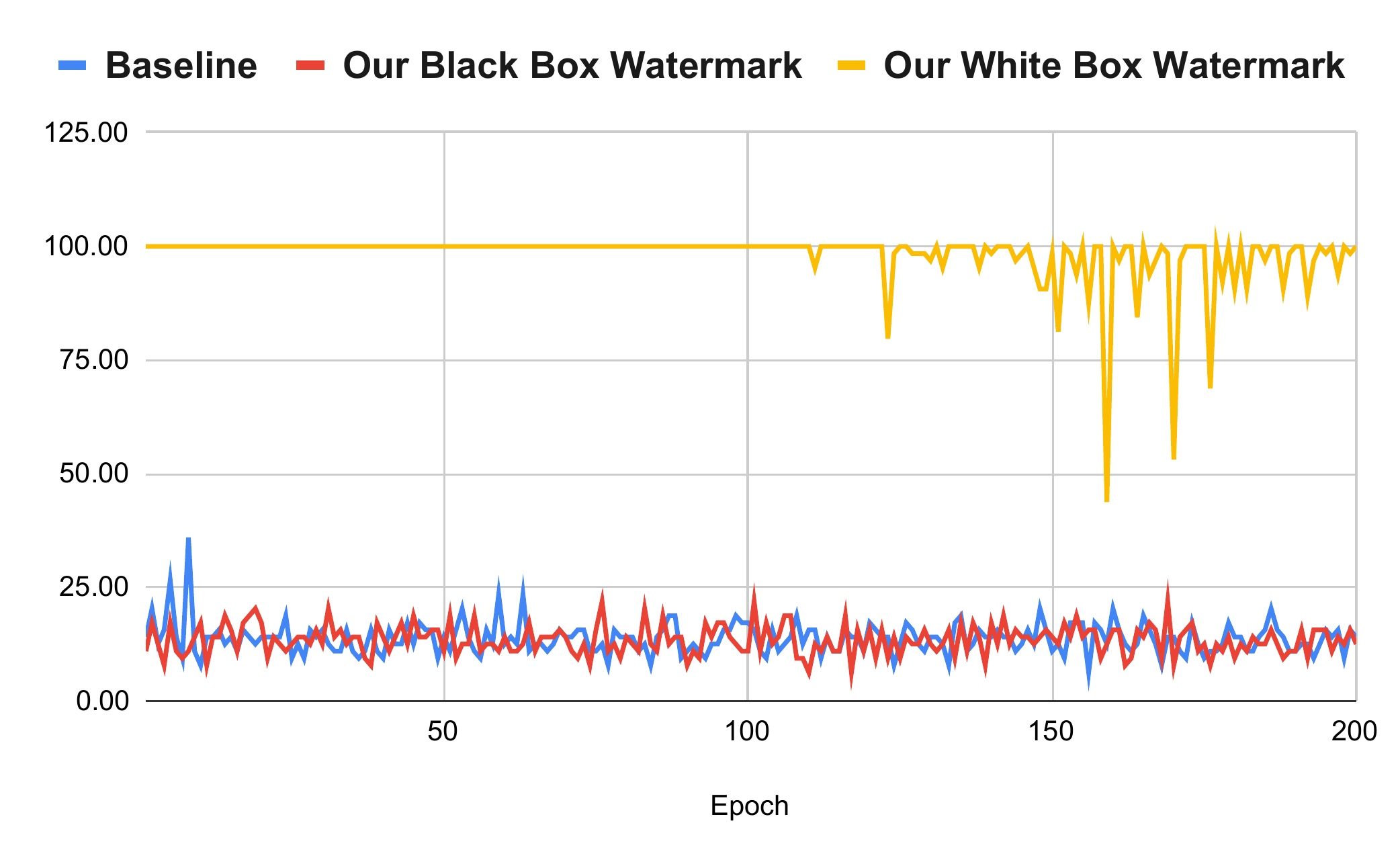}
         \caption{lr=.001}
     \end{subfigure}
     \hfill
    \begin{subfigure}[t]{0.45\textwidth}
         \centering
         \includegraphics[width=1\linewidth]{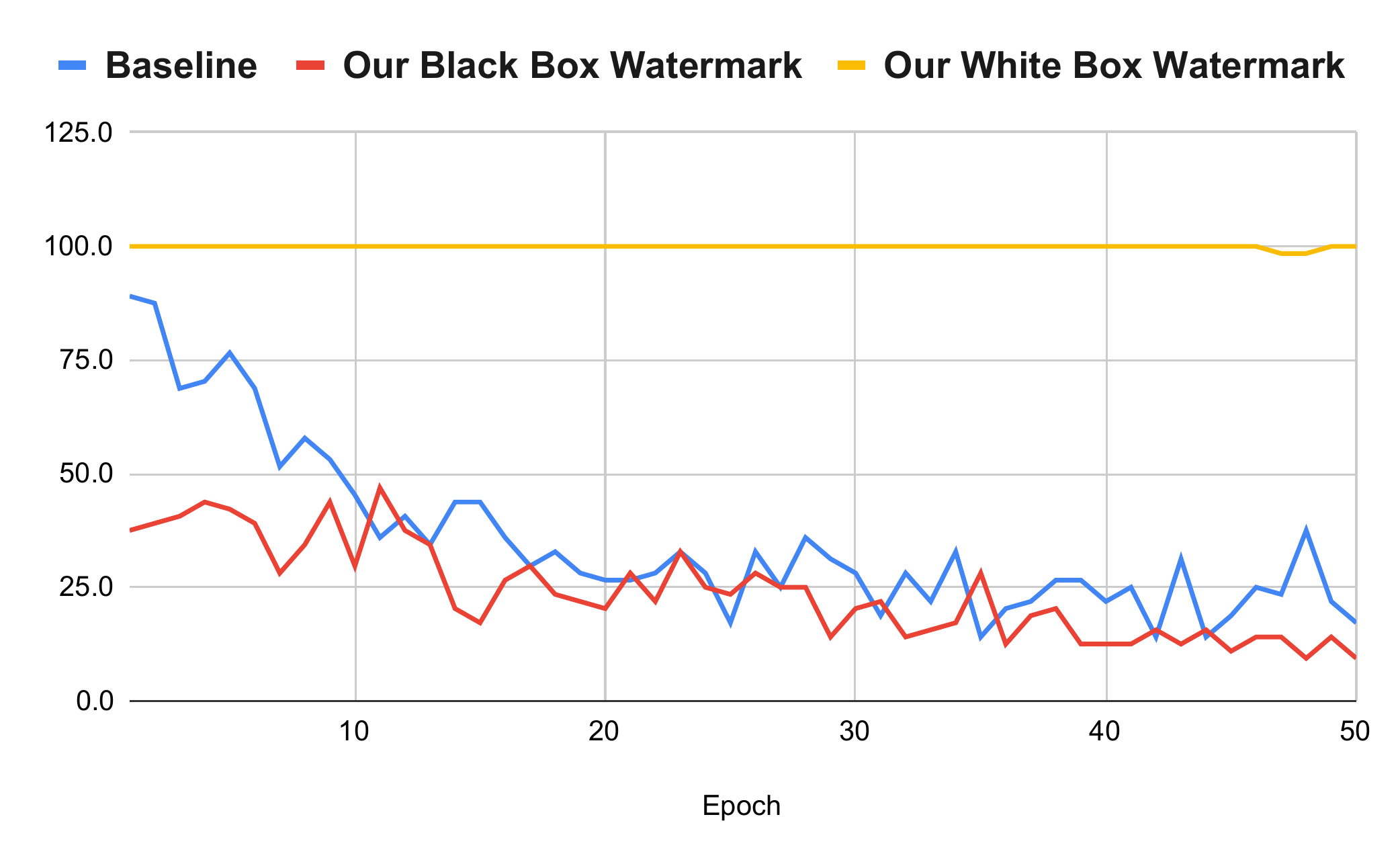}
         \caption{lr=.0001}
     \end{subfigure}
     \caption{CIFAR-10 trigger set accuracy when faced with finetuning attacks}
     \label{fig:cifar_ft}
\end{figure*}

\begin{figure*}[htb]
    \centering
    
    \begin{subfigure}[t]{0.45\textwidth}
         \centering
         \includegraphics[width=1\linewidth]{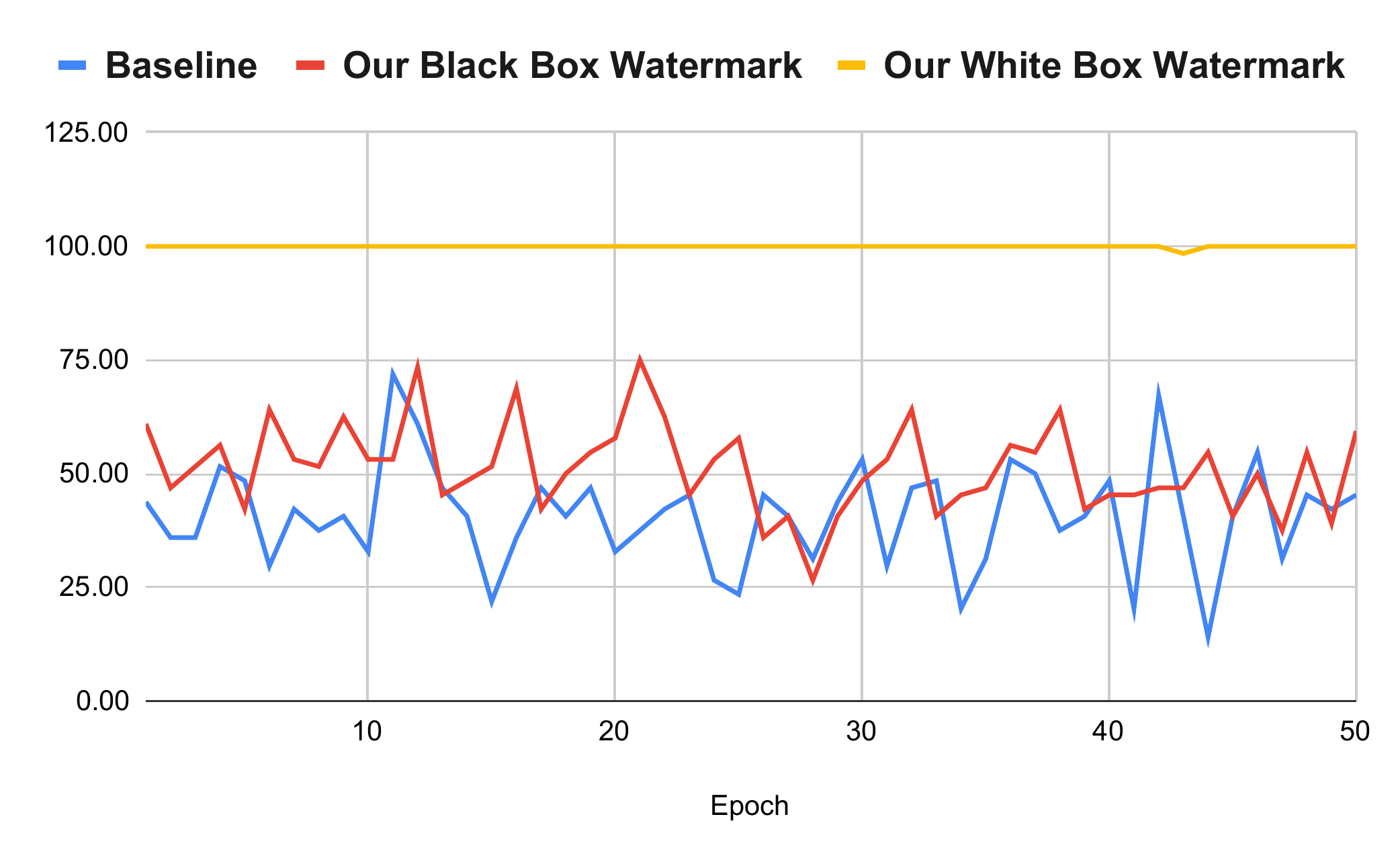}
         \caption{finetuning attack with lr=.001}
     \end{subfigure}
     \hfill
    \begin{subfigure}[t]{0.45\textwidth}
         \centering
         \includegraphics[width=1\linewidth]{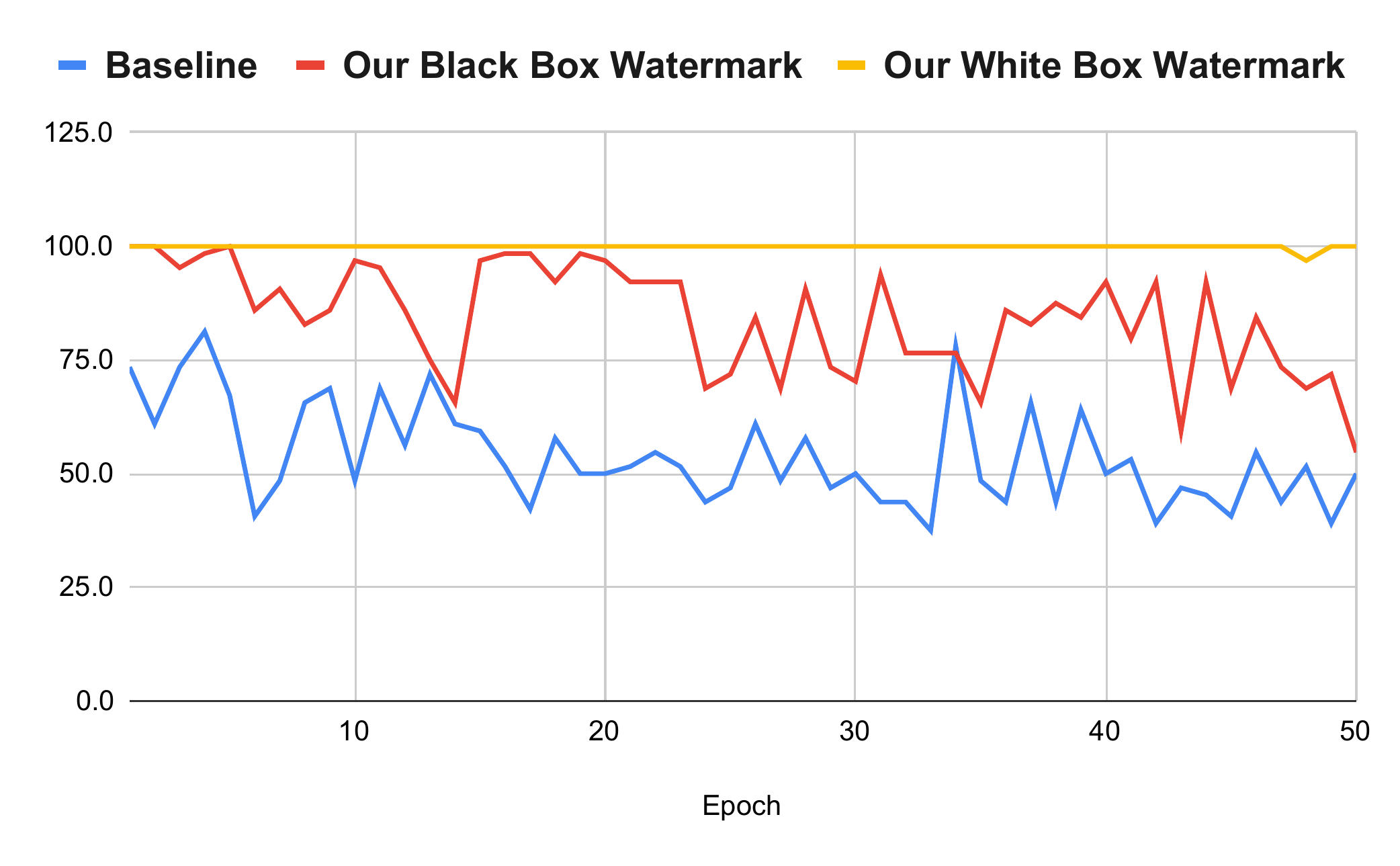}
         \caption{finetuning attack with lr=.0001}
     \end{subfigure}
     \caption{MNIST trigger set accuracy when faced with finetuning attacks}
     \label{fig:mnist_ft}
\end{figure*}

\begin{figure*}[htb]
    \centering
    \begin{subfigure}[t]{0.45\textwidth}
         \centering
         \includegraphics[width=1\linewidth]{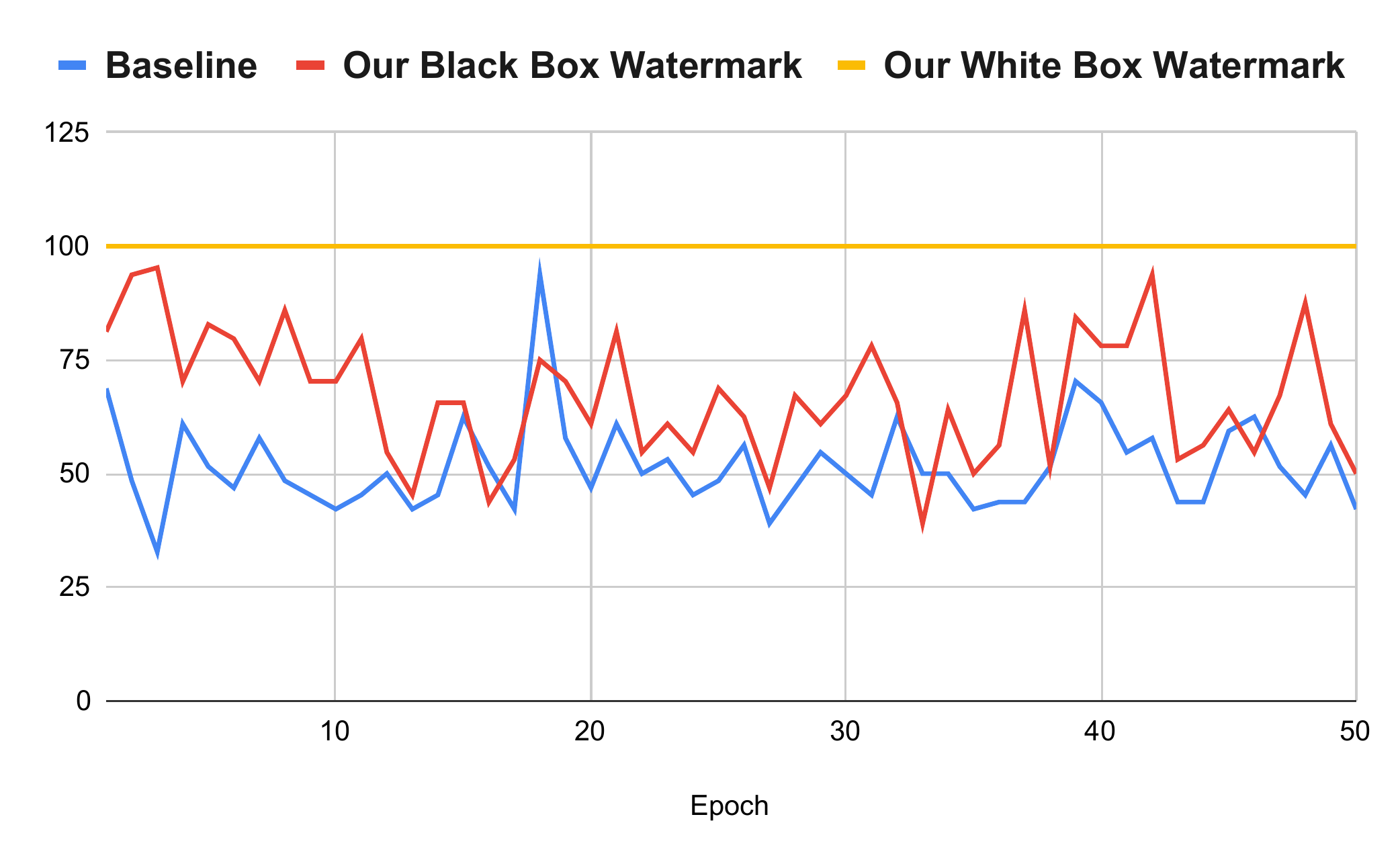}
         \caption{hard-label distillation with lr=1e-3}
     \end{subfigure}
     \hfill
    \begin{subfigure}[t]{0.45\textwidth}
         \centering
         \includegraphics[width=1\linewidth]{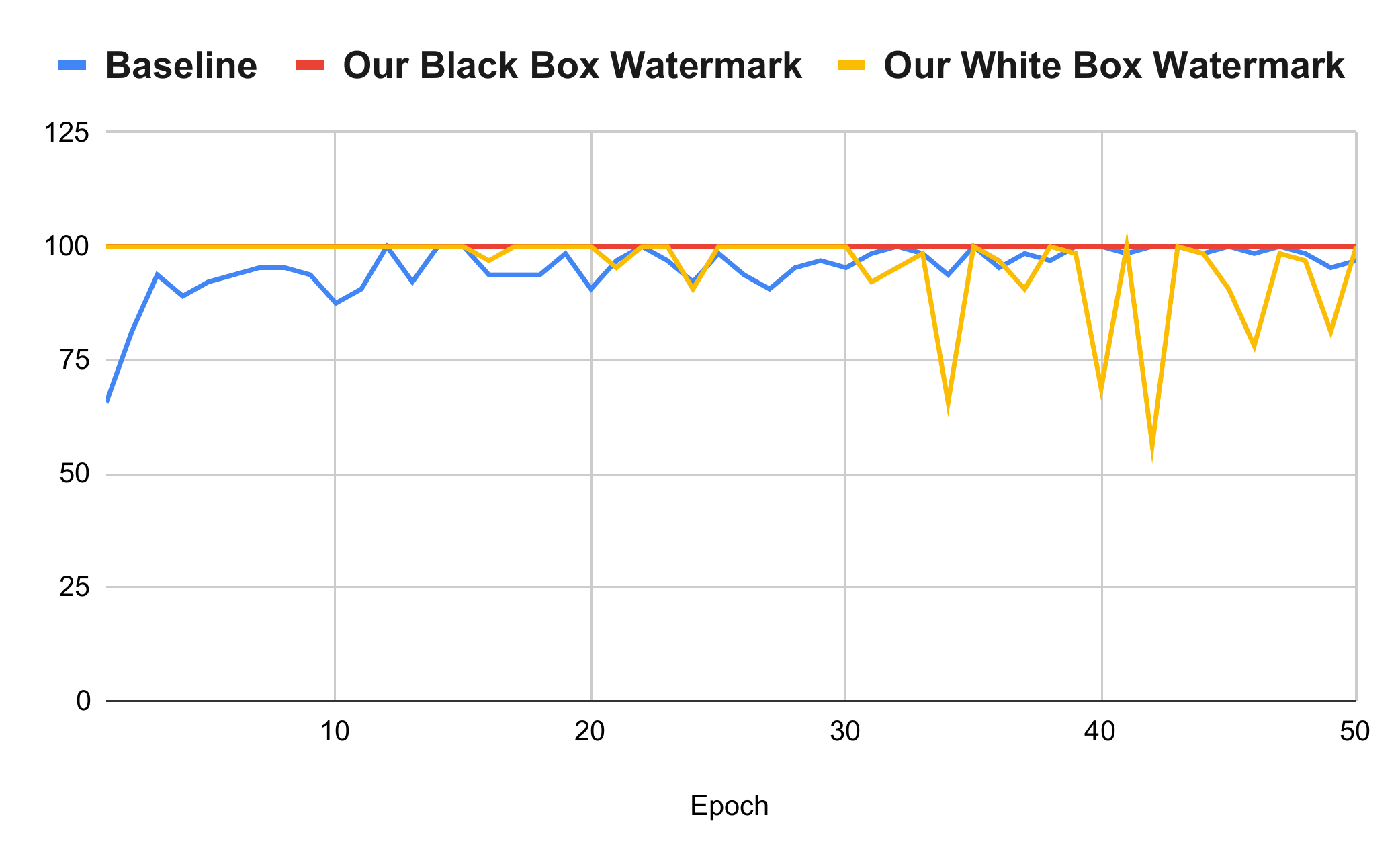}
         \caption{soft-label distillation with lr=1e-3}
     \end{subfigure}
     \caption{MNIST trigger set accuracy when faced with distillation attacks}
     \label{fig:mnist_d}
\end{figure*}

\begin{figure*}[htb]
    \centering
    \begin{subfigure}[t]{0.45\textwidth}
         \centering
         \includegraphics[width=1\linewidth]{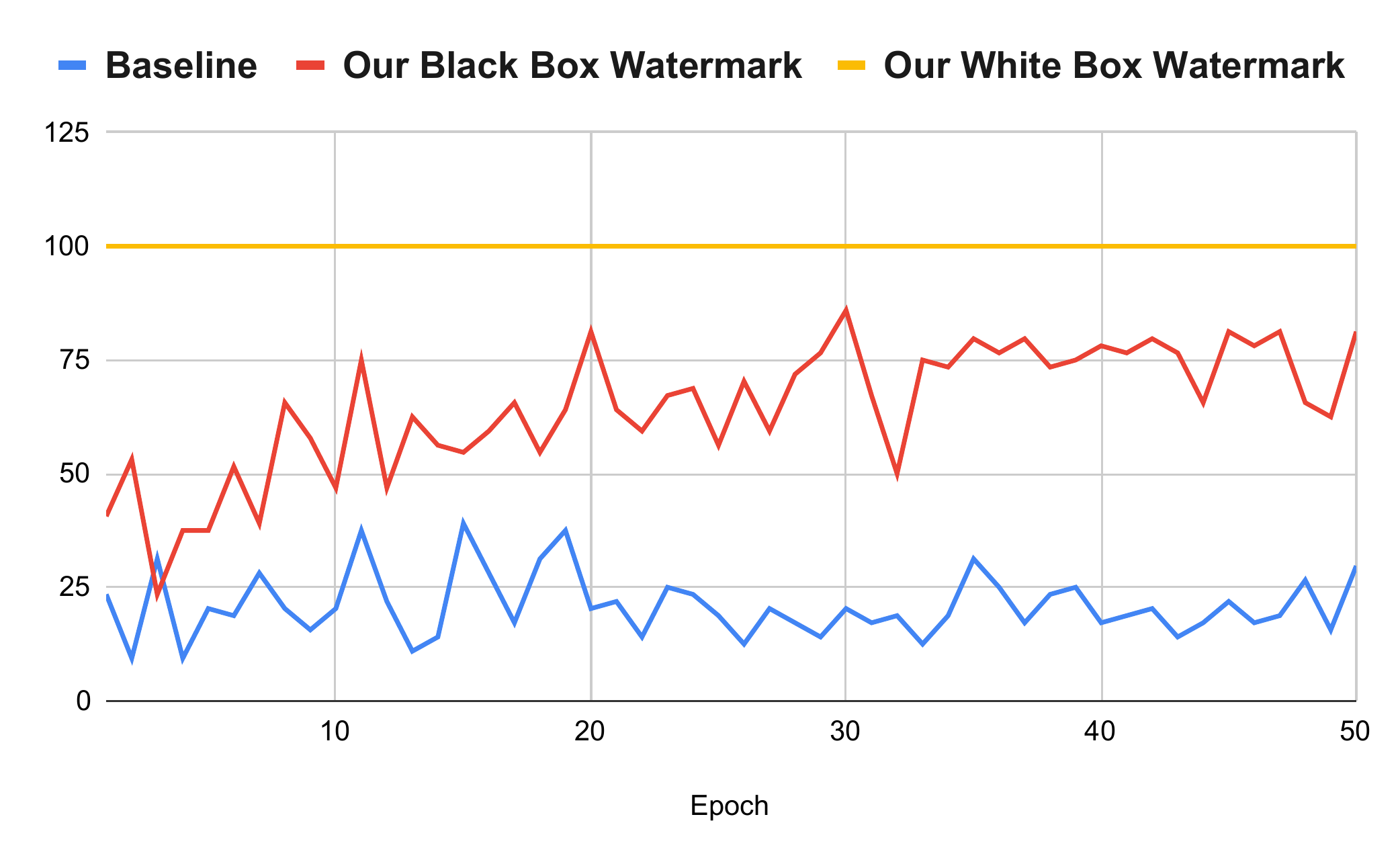}
         \caption{hard-label distillation with lr=1e-3}
     \end{subfigure}
     \hfill
    \begin{subfigure}[t]{0.45\textwidth}
         \centering
         \includegraphics[width=1\linewidth]{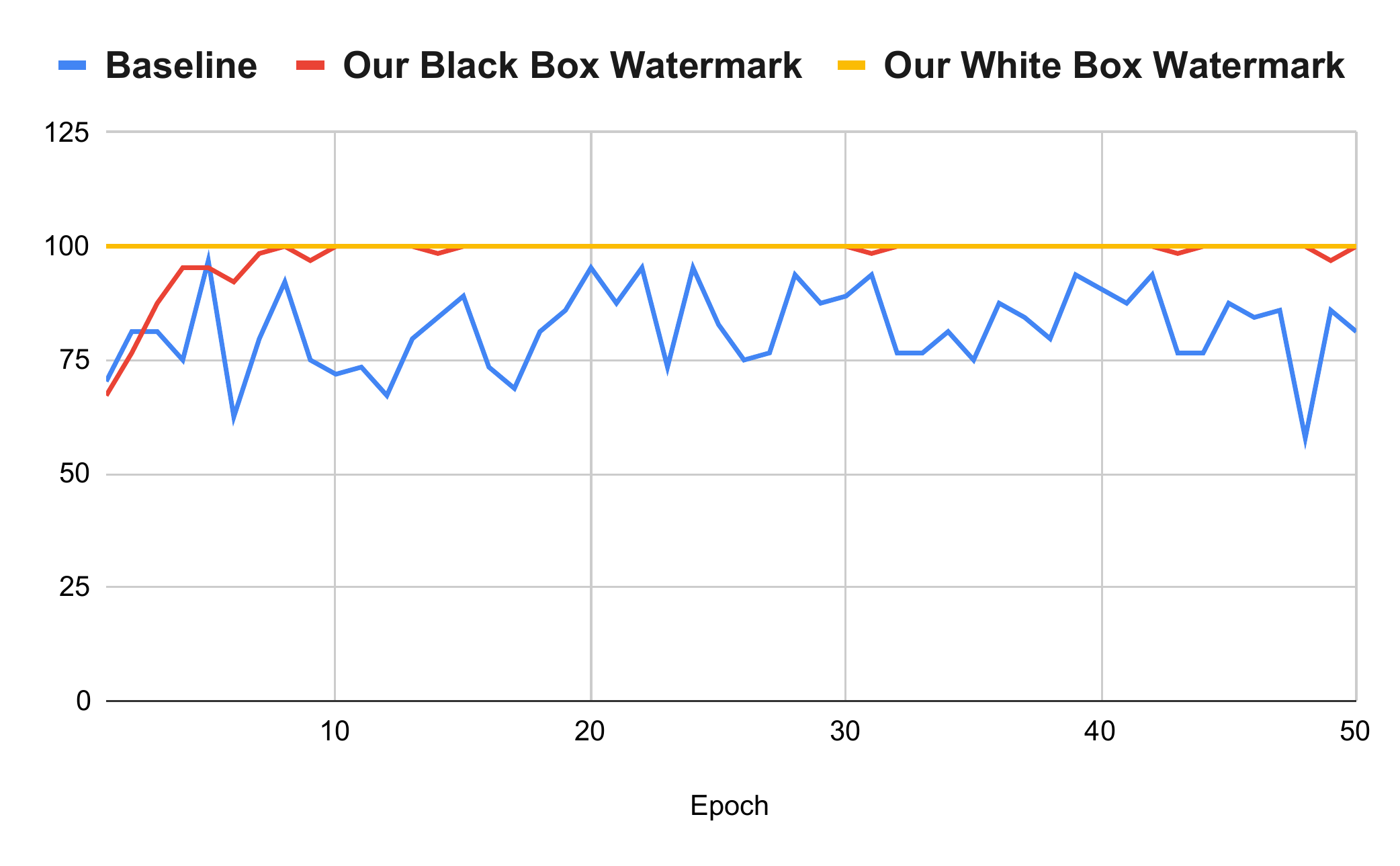}
         \caption{soft-label distillation with lr=1e-3}
     \end{subfigure}
     \caption{CIFAR-10 trigger set accuracy when faced with distillation attacks}
     \label{fig:cifar_d}
\end{figure*}

\begin{figure*}[htb]
    \centering
    \begin{subfigure}[t]{0.45\textwidth}
         \centering
         \includegraphics[width=1\linewidth]{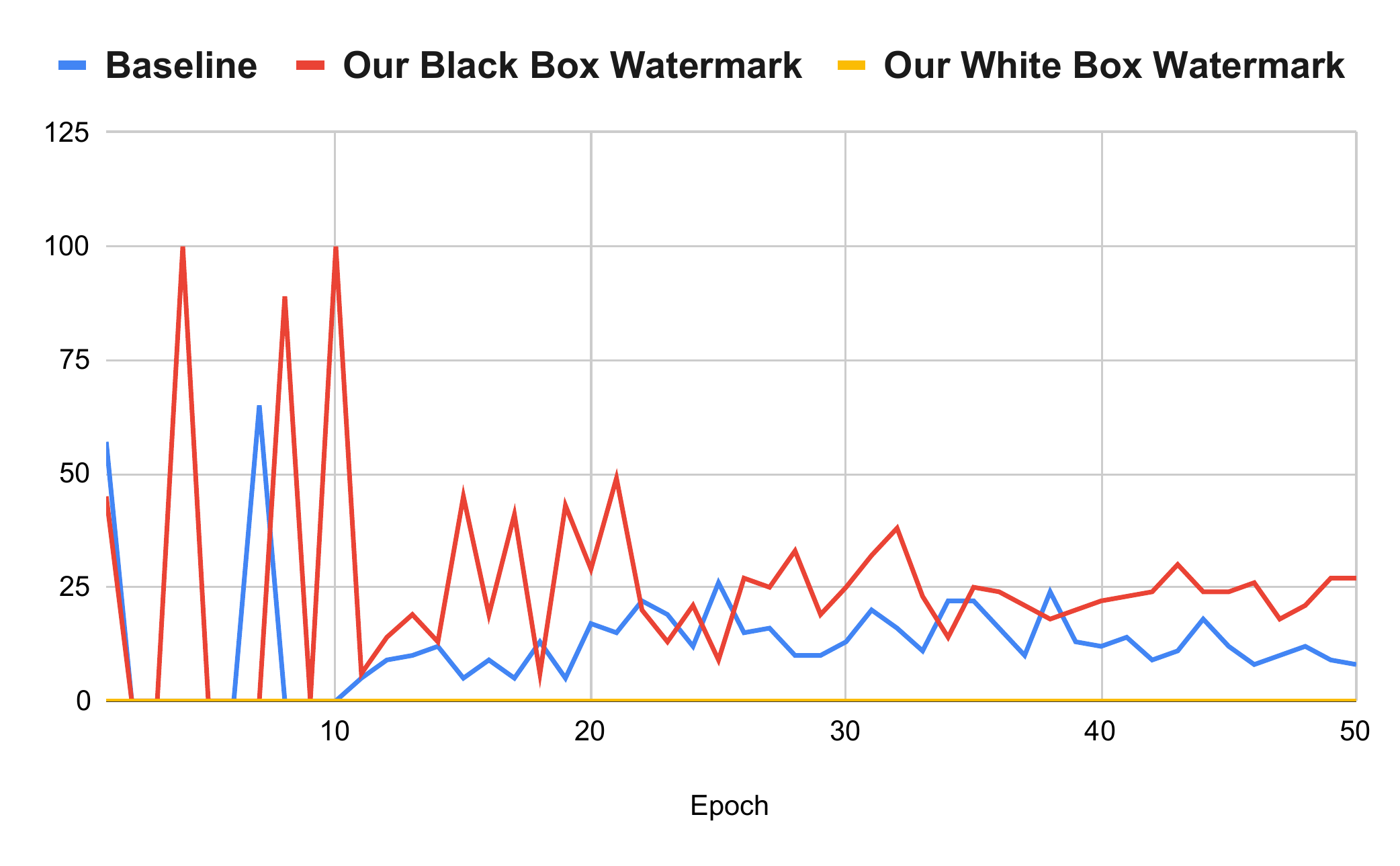}
         \caption{CIFAR-10}
     \end{subfigure}
     \hfill
    \begin{subfigure}[t]{0.45\textwidth}
         \centering
         \includegraphics[width=1\linewidth]{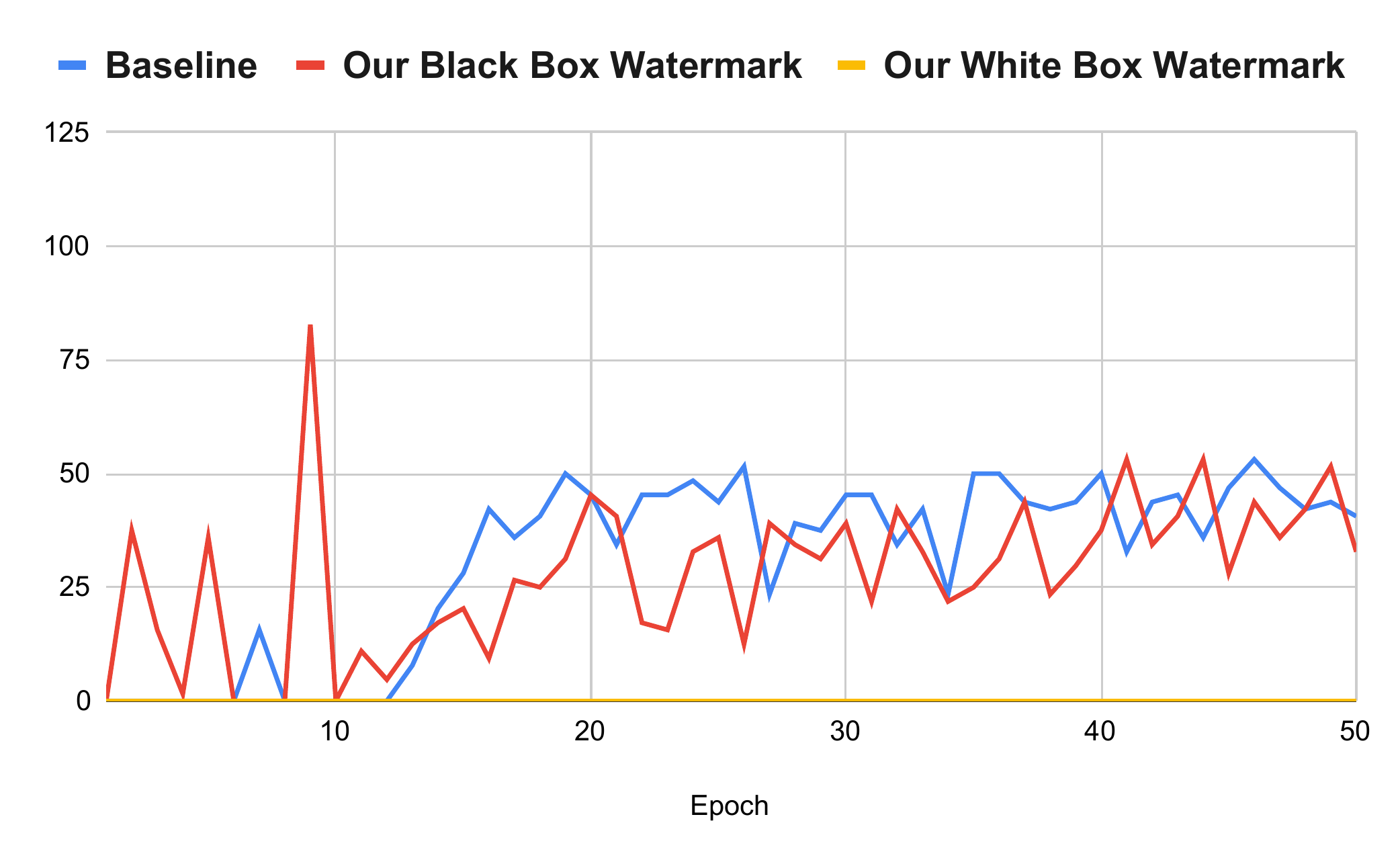}
         \caption{MNIST}
     \end{subfigure}
     \caption{Trigger set accuracy when faced with distillation+regularization attacks}
     \label{fig:distill+reg}
\end{figure*}

\clearpage

\subsection{Algorithm for empirical order statistic}

\begin{algorithm*}
\begin{algorithmic}
\Function{EmpiricalPercentile}{$n$, $c$, $\sigma$, $\epsilon$}
\State $p_{lower} \gets $ $\Phi(-\frac{\epsilon}{\sigma})$ \Comment{calculate theoretical percentile that we should be lower bounding}

\State $\hat{\underline{K}}_{lower}, \hat{\overline{K}}_{lower} \gets $ $0, \lfloor n \cdot p_{lower} \rfloor$ \Comment{initialized empirical order statistics for lower bound}

\While {$\hat{\overline{K}}_{lower} - \hat{\underline{K}}_{lower} > 1$}
\State $\dot{K}_{lower} \gets  \lfloor (\hat{\overline{K}}_{lower} + \hat{\underline{K}}_{lower})/2 \rfloor$
\If {1-Binomial(n, $\dot{K}_{lower}$,  $p_{lower}$) $> c$}
\State $\hat{\underline{K}}_{lower} \gets \dot{K}_{lower}$
\Else 
\State $\hat{\overline{K}}_{lower} \gets \dot{K}_{lower}$
\EndIf 
\EndWhile

\If {$\hat{\underline{K}}_{lower}>0$}
\State \textbf{return} $\hat{\underline{K}}_{lower}$
\Else
\State \textbf{return} null
\EndIf
\EndFunction

\end{algorithmic}
\caption{Choosing the empirical order statistics that sufficiently lower bound the theoretical percentile}
\label{alg:emp_order}
\end{algorithm*}
\clearpage
\subsection{$\ell_2$ norm change during attack}
\begin{table*}[thb]
\centering
\begin{tabular}{lllllllllll}
\toprule
Method &
  1st & 2 & 3 & 4 & 5 & 6 & 7 & 8 & 9 & 10\\
  \midrule
 CIFAR & & & & & & & & & & \\
 \midrule
 Hard label $10^{-4}$ &2.41&3.07&3.56&4.00&4.37&4.71&5.00&5.32&5.64&5.88 \\
  Hard label $10^{-3}$ &19.4&21.33&23.45&25.71&27.95&30.02&32.06&34.06&36.12&38.04\\
 Soft label $10^{-4}$ &2.06&2.47&2.73&2.95&3.2&3.47&3.73&3.97&4.16&4.38\\
  Soft label $10^{-3}$ &19.29&20.19&21.00&21.9&22.75&23.7&24.64&25.5&26.36&27.34 \\
 Finetune $10^{-4}$& 2.85&3.47&4.18&4.79&5.48&6.13&6.76&7.37&7.92&8.45\\
  Finetune $10^{-3}$& 19.93&22.57&25.54&28.41&31.34&34.31&37.31&40.18&42.98&45.73 \\
  \midrule
 MNIST & & & & & & & & & & \\
 \midrule
 Hard label $10^{-4}$ & 2.39&3.14&3.71&4.17&4.66&5.04&5.32&5.63&5.92&6.25 \\
  Hard label $10^{-3}$ & 17.58&19.34&21.2&22.87&24.77&26.73&28.77&30.33&32.12&33.83 \\
 Soft label $10^{-4}$ & 1.56&2.23&2.86&3.46&3.98&4.45&4.94&5.35&5.76&6.15\\
  Soft label $10^{-3}$& 20.35&22.51&25.00&28.12&30.29&32.31&34.35&36.58&38.84&41.1 \\
 Finetune $10^{-4}$ & 2.67&3.44&4.08&4.61&5.12&5.67&6.03&6.45&6.87&7.22\\
  Finetune $10^{-3}$ & 19.4&21.33&23.43&25.53&27.59&29.78&31.96&34.15&36.33&38.1\\

\bottomrule
\end{tabular}
\caption{Difference in $\ell_2$ norm from previous parameters after each epoch of attack. After the first epoch, the increase is general small on each successive epoch.}
\label{tab:l2movesfull}
\end{table*}

\subsection{Additional Persistence Evaluation}
\label{ap:persistence}
In this section, we evaluated our watermark scheme with respect to 11 more attacks from \cite{lukas2021sok}. We allow the adversary to have a time budget of 1 hour to remove the watermark as this is approximately the amount of time needed to train the model from scratch. With a budget any larger than 1 hour, the adversary will be better off training his/her own model. 

We consider a watermark removed if the adversary obtains a model with higher than 82\% test accuracy with less than 30\% of watermark accuracy. We follow conventions from \cite{lukas2021sok} where they consider an attack successful only if the accuracy of the model does not degrade by more than a certain amount and than the watermark accuracy remains above a decision threshold. We selected 82\% following the convention in \cite{lukas2021sok} where they consider an attack unsuccessful if the model's accuracy has been degraded by more than 5\%. On the other hand, we chose 30\% as a decision threshold as specified by \cite{lukas2021sok} to err on the conservative side.

We used the \href{https://github.com/dnn-security/Watermark-Robustness-Toolbox}{Watermark-Robustness-Toolbox} to conduct the additional persistence evaluation. For each of the attack, as discussed in the paper the defender has half of the original training dataset while the attacker has the other half. For different attacks discussed in \ref{tab:attacks}, we used the default hyper-parameters present in \href{https://github.com/dnn-security/Watermark-Robustness-Toolbox/tree/master/configs/cifar10/attack_configs}{/configs/cifar10/attack\_configs/} in which we simply changed the number of epochs to 100 in order to restrict the adversary within the time budget of approximately 1 hour. 

Within the time constraint, all the methods tested fail to remove both the black-box watermark and white-box watermark simultaneously. Even though neural cleanse and neural laundering are showing some effects at removing the watermark, these two methods did not successfully remove the watermark within the time limit. The two approaches also result in greater loss of test accuracy. The fine-tuning based approach (FTAL FTLL, RTAL, and RTLL) surprisingly increases the test accuracy. This is consistent with the results in \cite{lukas2021sok} where the finetuning based approaches increase the test accuracy due to the use of additional training data.

\begin{table}[h]
\centering
\scalebox{1}{
\begin{tabular}{lccccc}
\toprule
 & \multicolumn{1}{l}{\begin{tabular}[c]{@{}l@{}}BB WM \\ removal time\end{tabular}} & \multicolumn{1}{l}{BB Acc} & \multicolumn{1}{l}{\begin{tabular}[c]{@{}l@{}}WB WM\\ removal time\end{tabular}} & \multicolumn{1}{l}{WB Acc} & \multicolumn{1}{l}{\begin{tabular}[c]{@{}l@{}}Accuracy\\ Loss \end{tabular}} \\
 \midrule
FTAL & 99 & 15 & 3600+ & 96 & -4.01 \\
FTLL & 3600+ & 62 & 813 & 0 & -2.24 \\
RTAL & 3600+ & 56 & 34 & 0 & -1.22 \\
RTLL & 3600+ & 100 & 32 & 0 & -0.33 \\
Adversarial Training & 530 & 28 & 3600+ & 100 & 3.72 \\
Neural Cleanse & 3600+ & 47 & 3600+ & 100 & 0.88 \\
Neural Laundering & 3600+ & 55 & 3600+ & 82 & 2.53 \\
Weight Quantization & 3600+ & 46 & 3600+ & 100 & 1.83 \\
Feature Shuffling & 0.97 & 100 & - & 100 & 0.05 \\
Weight Pruning & 3.27 & 100 & - & 100 & 0.05 \\
Weight Shifting & 3600+ & 52 & 3600+ & 100 & 2.49 \\
\bottomrule
\end{tabular}
}
\caption{Evaluation against most attacks with new metrics}
\label{tab:attacks}
\end{table}

\end{document}